%% file: main.tex
\documentclass[11pt,a4paper]{article}
\usepackage{acl2020}
\usepackage{latexsym}

\usepackage{microtype}
\usepackage{mdframed}
\usepackage{colortbl}
\usepackage{etoolbox}

\input{math-shortcuts}

\aclfinalcopy
\def\aclpaperid{158}
\graphicspath{{../figures/}{../}}
\graphicspath{{../qualitative/}{../}}

\definecolor{pink}{rgb}{0.858, 0.188, 0.478}
\definecolor{darkviolet}{rgb}{0.293, 0.0, 0.508}

\graphicspath{{../figures/}{../}}
\graphicspath{{../submission/}{../}}

\newrobustcmd*{\bftabnum}{%
  \bfseries
  \sisetup{output-decimal-marker={\textmd{.}}}%
}

\usepackage{xpatch}
\makeatletter
\AtBeginDocument{\xpatchcmd{\@thm}{\thm@headpunct{.}}{\thm@headpunct{}}{}{}}
\makeatother

\newcommand\rtwor{\textsc{r2r}\xspace}
\newcommand\RtwoR{\textsc{Room2Room}\xspace}
\newcommand\rfourr{\textsc{r4r}\xspace}
\newcommand\RfourR{\textsc{Room4Room}\xspace}
\newcommand\rsixr{\textsc{r6r}\xspace}
\newcommand\RsixR{\textsc{Room6Room}\xspace}
\newcommand\reightr{\textsc{r8r}\xspace}
\newcommand\ReightR{\textsc{Room8Room}\xspace}
\newcommand\bbwalk{\textsc{BabyWalk}\xspace}
\newcommand\bbstep{\textsc{Baby-Step}\xspace}
\newcommand\bbsteps{\textsc{Baby-Step}s\xspace}

\newcommand\eat[1]{}

\newcommand\mypara[1]{\paragraph{#1}}

\newcommand{\SM}{{Appendix}\xspace}
\newcommand{\summary}{{f_{\textsc{summary}}}}
\newcommand{\ourmethod}{\bbwalk}
\newcommand{\ourtitle}{{BabyWalk: Going Farther in Vision-and-Language Navigation\\by Taking Baby Steps}}

\title{\ourtitle}

\author{
\textbf{Wang Zhu}\thanks{\xspace\xspace Author contributed equally}\xspace\xspace$^{1}$ 
\quad \textbf{Hexiang Hu}\textcolor{red}{\footnotemark[1]}\xspace\xspace$^{2}$ 
\quad \textbf{Jiacheng Chen}$^{2}$ 
\quad \textbf{Zhiwei Deng}$^3$ \\[4pt]
\quad \textbf{Vihan Jain}$^{4}$ 
\quad \textbf{Eugene Ie}$^{4}$
\quad \textbf{Fei Sha}\thanks{\xspace\xspace On leave from University of Southern California}\xspace\xspace$^{2, 4}$ \\[6pt]
$^1$Simon Fraser University
\quad $^2$University of Southern California
\quad $^3$Princeton University \\[4pt]
\quad $^4$Google Research
}

\begin{document}
\maketitle

\begin{abstract}
Learning to follow instructions is of fundamental importance to autonomous agents for vision-and-language navigation (VLN). In this paper, we study how an agent can navigate long paths when learning from a corpus that consists of shorter ones.  We show that existing state-of-the-art agents do not generalize well. To this end, we propose BabyWalk, a new VLN agent that is learned to navigate by decomposing long instructions into shorter ones (BabySteps) and completing them sequentially. A special design memory buffer is used by the agent to turn its past experiences into contexts for future steps. The learning process is composed of two phases. In the first phase, the agent uses imitation learning from demonstration to accomplish BabySteps. In the second phase, the agent uses curriculum-based reinforcement learning to maximize rewards on navigation tasks with increasingly longer instructions. We create two new benchmark datasets (of long navigation tasks) and use them in conjunction with existing ones to examine BabyWalk's generalization ability. Empirical results show that BabyWalk achieves \textit{state-of-the-art} results on several metrics, in particular, is able to follow long instructions better.  The codes and the datasets are released on our project page \url{https://github.com/Sha-Lab/babywalk}.
\end{abstract}

\input{intro}
\input{related}

\input{problem}

\input{method}

\input{exp}

\input{discussion}

\mypara{Acknowledgments} {\small
We appreciate the feedback from the reviewers. 
This work is partially supported by NSF Awards IIS-1513966/1632803/1833137, CCF-1139148, DARPA Award\#: FA8750-18-2-0117, 
DARPA-D3M - Award UCB-00009528, Google Research Awards, 
gifts from Facebook and Netflix, and ARO\# W911NF-12-1-0241 and W911NF-15-1-0484.
}

{
\bibliography{main}
\bibliographystyle{acl_natbib}
}

\clearpage
\appendix

{
    \centering
    \textbf{\large Appendix} \\[2pt]
}
\input{supp_content}

\end{document}

%% file: math-shortcuts.tex
\usepackage{times}
\usepackage{epsfig}
\usepackage{graphicx}
\usepackage{float}
\usepackage{wrapfig}
\usepackage{amsmath,amssymb,amsthm}
\usepackage{algorithm,algorithmicx,algpseudocode}
\usepackage{bm,xspace}
\usepackage{comment}
\usepackage{verbatim}
\usepackage{multirow}
\usepackage{array}
\usepackage{balance}
\usepackage{url}
\usepackage{booktabs}
\usepackage{etoolbox,siunitx}
\usepackage{calc}
\usepackage{pifont,hologo}
 
\usepackage{nicefrac}

\usepackage{arydshln}
\usepackage[utf8]{inputenc}
\usepackage[T1]{fontenc}
\usepackage{url} 
\usepackage{amsfonts}
\usepackage{nicefrac}
\usepackage{microtype}
\usepackage[american]{babel}
\usepackage{enumitem}
\usepackage{amssymb}
\usepackage{amsmath}
\usepackage{caption}
\usepackage{subcaption}
\usepackage{siunitx}

\usepackage[super]{nth}
\usepackage{nicefrac}
\robustify\bfseries

\usepackage{dsfont}

\def\bspsi{\boldsymbol{\psi}}
\def\bspi{\boldsymbol{\pi}}

\def\bsa{\boldsymbol{a}}

\def\bso{\boldsymbol{o}}

\def\bss{\boldsymbol{s}}

\def\bsx{\boldsymbol{x}}
\def\bsy{\boldsymbol{y}}
\def\bsz{\boldsymbol{z}}

\def\rmA{\mathrm{A}}

\def\rmM{\mathrm{M}}

\def\rmT{\mathrm{T}}

\def\rmX{\mathrm{X}}
\def\rmY{\mathrm{Y}}

\def\calA{\mathcal{A}}

\def\calO{\mathcal{O}}

\def\calS{\mathcal{S}}

\def\bbR{\mathbb{R}}

\DeclareMathSymbol{@}{\mathord}{letters}{"3B}

\newcommand\paren[1]{\left(#1\right)}
\newcommand\braces[1]{\left\{#1\right\}}
\newcommand\brackets[1]{\left[#1\right]}
\newcommand\abs[1]{\left\lvert#1\right\rvert}

\makeatletter
\DeclareRobustCommand\onedot{\futurelet\@let@token\@onedot}
\def\@onedot{\ifx\@let@token.\else.\null\fi\xspace}

\def\ie{\emph{i.e}\onedot} 
\def\cf{\emph{c.f}\onedot} 
\def\etc{\emph{etc}\onedot}

\makeatother

%% file: intro.tex
% !TEX root = main.tex

\section{Introduction}
\label{sec:intro}

Autonomous agents such as household robots need to interact with the physical world in multiple modalities. As an example, in vision-and-language navigation (VLN)~\citep{anderson2018vision}, the agent moves around in a photo-realistic simulated environment~\citep{Matterport3D} by following a sequence of natural language instructions. To infer its whereabouts so as to decide its moves, the agent infuses its visual perception, its trajectory and the instructions~\citep{fried2018speaker,anderson2018vision,wang2019reinforced,ma2019self,ma2019regretful}.

Arguably, the ability to understand and follow the instructions is one of the most crucial skills to acquire by VLN agents. \citet{jain2019stay} shows that the VLN agents trained on the originally proposed dataset \RtwoR (\ie\rtwor thereafter) do \emph{not} follow the instructions, despite having achieved high success rates of reaching the navigation goals. They proposed two remedies: a new dataset \RfourR (or \rfourr) that doubles the path lengths in the \rtwor, and a new evaluation metric Coverage weighted by Length Score (CLS) that measures more closely whether the ground-truth paths are followed. They showed  optimizing the fidelity of following instructions leads to agents with desirable behavior. Moreover, the long lengths in \rfourr are informative in identifying agents who score higher in such fidelity measure.

In this paper, we investigate another crucial aspect of following the instructions: \emph{can a VLN agent generalize to following longer instructions by learning from shorter ones?}  This aspect has important implication to real-world applications as collecting annotated long sequences of instructions and training on them can be costly. Thus,  it is highly desirable to have this generalization ability. After all, it seems that humans can achieve this effortlessly\footnote{Anecdotally, we do not have to learn from long navigation experiences. Instead, we extrapolate from our experiences of learning to navigate in shorter distances or smaller spaces (perhaps a skill we learn when we were babies or kids).}.

\begin{figure}[t]
\centering
\includegraphics[width=0.45\textwidth]{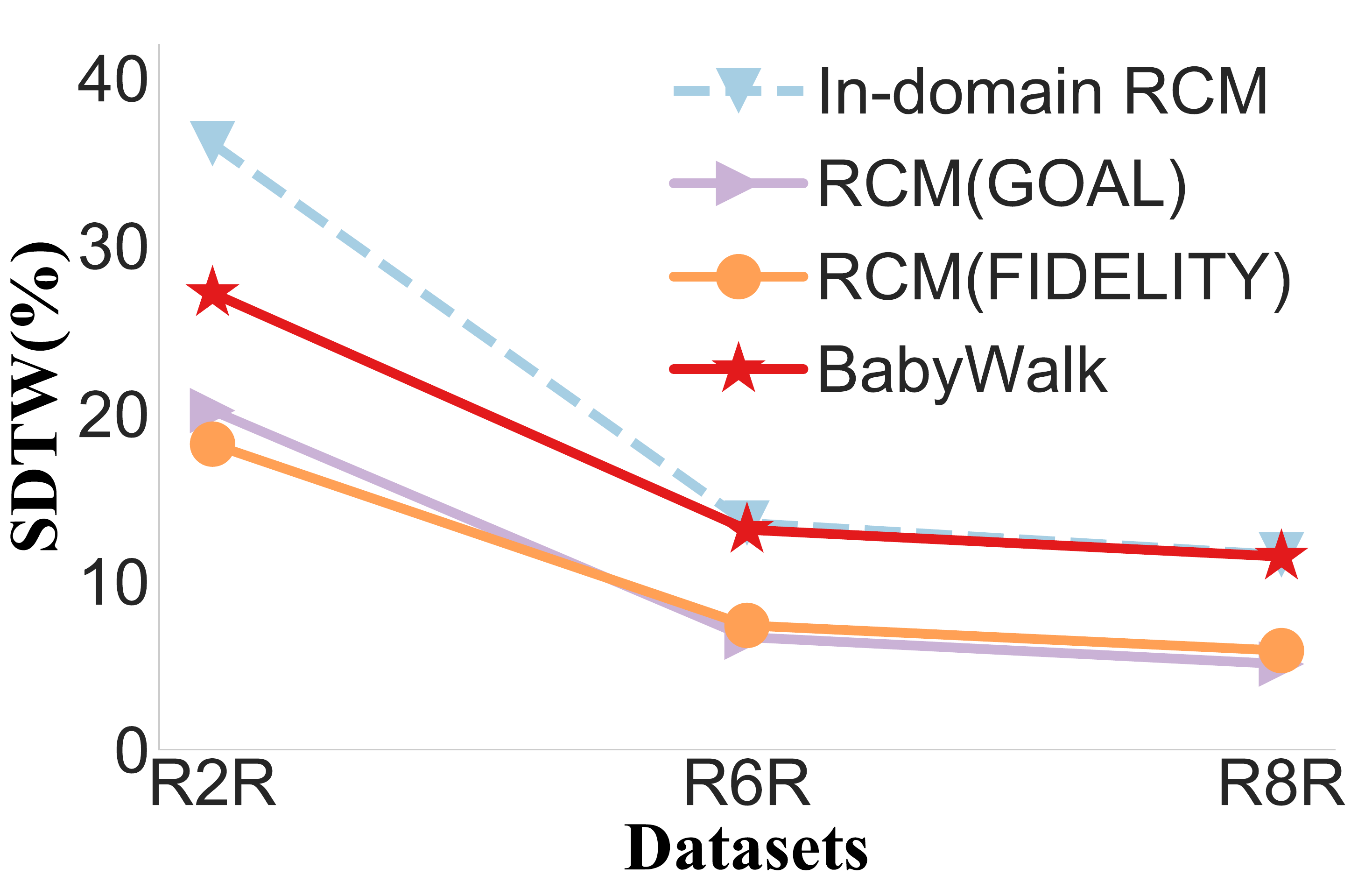}
\caption{Performance of various VLN agents on generalizing from shorter navigation tasks to longer ones. The vertical axis is the newly proposed path-following metric SDTW~\citep{magalhaes2019effective}, the higher the better. \bbwalk generalizes better than other approaches across different lengths of navigation tasks. Meanwhile, it get very close to the performances of the in-domain agents (the dashed line). Please refer to the texts for details. }
\label{fig:teaser}
\end{figure}

To this end, we have created several datasets of longer navigation tasks, inspired by \rfourr~\citep{jain2019stay}. We trained VLN agents on \rfourr and use the agents to navigate in \RsixR (\ie, \rsixr) and \ReightR (\ie, \reightr). We contrast to the performance of the agents which are trained on those datasets directly (``in-domain''). The results are shown in Fig.~\ref{fig:teaser}.

Our findings are that the agents trained on \rfourr (denoted by the purple and the pink solid lines) perform significantly worse than the \emph{in-domain} agents (denoted the light blue dashed line). Also interestingly, when such  \emph{out-of-domain} agents are applied to the dataset \rtwor with \emph{shorter} navigation tasks, they also perform significantly worse than the corresponding in-domain agent despite \rfourr containing many navigation paths from \rtwor. Note that the agent trained to optimize the aforementioned fidelity measure (\textsc{rcm}(fidelity)) performs better than the  agent trained to reach the goal only (\textsc{rcm}(goal)), supporting the claim by \citet{jain2019stay} that following instructions is a more meaningful objective than merely goal-reaching. Yet, the fidelity measure itself is not enough to enable the agent to transfer well to longer navigation tasks.

To address these deficiencies, we propose a new approach for VLN. The agent follows a long navigation instruction by decomposing the instruction into shorter ones (``micro-instructions'', \ie, \bbsteps), each of which corresponds to an intermediate goal/task to be executed sequentially. To this end, the agent has three components: (a) a memory buffer that summarizes the agent's experiences  so that the agent can use them to provide the context for executing the next \bbstep. (b) the agent first learns from human experts in ``bite-size''. Instead of trying to imitate to achieve the ground-truth paths as a whole, the agent is given the pairs of  a \bbstep and the corresponding human expert path so that it can learn policies of actions from shorter instructions. (c) In the second stage of learning, the agent refines the policies by curriculum-based reinforcement learning, where the agent is given increasingly longer navigation tasks to achieve.  In particular, this curriculum design reflects our desiderata that the agent optimized on shorter tasks should generalize well to slightly longer tasks and then much  longer ones. 

While we do not claim that our approach faithfully simulates human learning of navigation, the design is loosely inspired by it. We name our approach \bbwalk and refer to the intermediate navigation goals in (b) as \bbsteps. Fig.~\ref{fig:teaser} shows that \bbwalk (the red solid line) significantly outperforms other approaches and despite being out-of-domain, it even reach the performance of in-domain agents on \rsixr and \reightr.

The effectiveness of \bbwalk also leads to an interesting twist. As mentioned before, one of the most important observations by \citet{jain2019stay} is that the original VLN dataset \rtwor fails to reveal the difference between optimizing goal-reaching (thus ignoring the instructions)  and optimizing the fidelity (thus adhering to the instructions). Yet, leaving details to section~\ref{sec:experiment}, we have also shown that applying \bbwalk to \rtwor can lead to equally strong  performance on generalizing from shorter instructions (\ie, \rtwor) to longer ones. 

In summary, in this paper, we have demonstrated empirically that the current VLN agents are ineffective in generalizing from learning on shorter navigation tasks to longer ones. We propose a new approach in addressing this important problem. We validate the approach with extensive benchmarks, including ablation studies to identify the effectiveness of various components in our approach. 

%% file: related.tex
% !TEX root = main.tex
\section{Related Work}
\label{sec:related}

\mypara{Vision-and-Language Navigation (VLN)} Recent works~\citep{anderson2018vision,thomason2019vision,jain2019stay,chen2019touchdown,nguyen2019help} extend the early works of instruction based navigation~\citep{chen2011learning,kim2013adapting,mei2016listen} to photo-realistic simulated environments. For instance, \citet{anderson2018vision} proposed to learn a multi-modal Sequence-to-Sequence agent (Seq2Seq) by imitating expert demonstration.
\citet{fried2018speaker}  developed a method that augments the paired instruction and demonstration data using a learned speaker model, to teach the navigation agent to better understand instructions. 
~\citet{wang2019reinforced} further applies reinforcement learning (RL) and self-imitation learning to improve navigation agents. \citet{ma2019self,ma2019regretful} designed models that track the execution progress for a sequence of instructions using soft-attention. 

Different from them, we focus on transferring an agent's performances on shorter tasks to longer ones. This leads to designs and learning schemes that improve generalization across datasets. We use a memory buffer to prevent mistakes in the distant past from exerting strong influence on the present.  In imitation learning stage, we solve fine-grained subtasks (\bbsteps) instead of asking the agent to learn the navigation trajectory as a whole. We then use curriculum-based reinforcement learning by asking the agent to follow increasingly longer instructions. 

\mypara{Transfer and Cross-domain Adaptation} There have been a large body of works in transfer learning and generalization across tasks and environments in both computer vision and reinforcement learning~\cite{andreas2017modular,oh2017zero,zhu2017visual,zhu2017target,sohn2018hierarchical,hu2018synthesized}. Of particular relevance is the recent work on adapting VLN agents to changes in visual environments~\citep{huang2019transferable,tan2019learning}. To our best knowledge, this work is the first to focus on adapting to a simple aspect of language variability --- the length of the instructions. 

\mypara{Curriculum Learning} Since proposed in~\citep{bengio2009curriculum}, curriculum learning was successfully used in a range of tasks:  training robots for goal reaching~\citep{florensa2017reverse}, visual question answering~\citep{mao2019neuro}, image generation~\citep{karras2017progressive}. To our best knowledge, this work is the first to apply the idea to learning in VLN.

%% file: problem.tex
% !TEX root = main.tex
\section{Notation and the Setup of VLN}
\label{sec:problem}

In the VLN task, the agent receives a natural language instruction $\rmX$ composed of a sequence of sentences. We model the agent with an \emph{Markov Decision Process} (MDP) which is defined as a tuple of a state space $\calS$, an action space $\calA$, an initial state $\bss_1$, a stationary transition dynamics $\rho: \calS \times \calA \to \calS$, a reward function $r: \calS \times \calA \to \bbR$, and the discount factor $\gamma$ for weighting future rewards.  The agent acts according to a policy $\bspi: \calS \times \calA \to 0 \cup R^+$. The state and action spaces are defined the same as in~\cite{fried2018speaker} (cf. \S~\ref{subsec:impl} for details).

For each $\rmX$, the sequence of the pairs $(\bss, \bsa)$ is called a trajectory $\rmY = \braces{\bss_1, \bsa_1, \ldots, \bss_{\abs{\rmY}}, \bsa_{\abs{\rmY}}}$ where $\abs{\cdot}$ denotes the length of the sequence or the size of a set. We use $\hat{\bsa}$ to denote an action taken by the agent according to its policy. Hence, $\hat{\rmY}$ denotes the agent's trajectory, while $\rmY$ (or $\bsa$)  denotes the human expert's trajectory (or action). The agent is given training examples of  $(\rmX, \rmY)$ to optimize its policy to maximize its expected rewards.

In our work, we introduce additional notations in the following. We will segment a (long) instruction $\rmX$ into multiple shorter sequences of sentences $\{\bsx_m, m = 1, 2, \cdots, \rmM\}$, to which we refer as \bbsteps. Each $\bsx_m$  is interpreted as a micro-instruction that corresponds to a trajectory by the agent $\hat{\bsy}_m$ and is aligned with a part of the human expert's trajectory, denoted as $\bsy_m$.  While the alignment is not available in existing datasets for VLN, we will describe how to obtain them in a later section (\S~\ref{sec:dataset}). Throughout the paper, we also freely interexchange the term ``following the $m$th micro-instruction'', ``executing the \bbstep $\bsx_m$'', or ``complete the $m$th subtask''.

We use $t\in[1, \abs{\rmY}]$ to denote the (discrete) time steps the agent takes actions. Additionally, when the agent follows $\bsx_m$, for convenience, we sometimes use $t_m \in [1, \abs{\hat{\bsy}_m}]$ to index the time steps, instead of the ``global time'' $t = t_m + \sum_{i=1}^{m-1} \abs{\hat{\bsy}_i}$.

%% file: method.tex
% !TEX root = main.tex
\section{Approach}
\label{sec:method}

We describe  in detail the 3 key elements in the design of our navigation agent: (i) a memory buffer for storing and recalling past experiences to provide contexts for the current navigation instruction (\S~\ref{subsec:agent}); (ii) an imitation-learning stage of navigating with short instructions to accomplish a single \bbstep (\S~\ref{subsec:imitation}); (iii) a curriculum-based reinforcement learning phase where the agent learns with increasingly longer instructions (\ie multiple \bbsteps) (\S~\ref{subsec:reinforce}). We describe new benchmarks created for learning and evaluation and key implementation details in \S~\ref{sec:dataset} and \S~\ref{subsec:impl} (with more details in the \SM).

\input{method_model}
\input{method_algo}

\input{method_dataset}
\input{method_impl}

%% file: method_model.tex
% !TEX root = main.tex

\begin{figure}[t]
    \centering
    \includegraphics[width=0.47\textwidth]{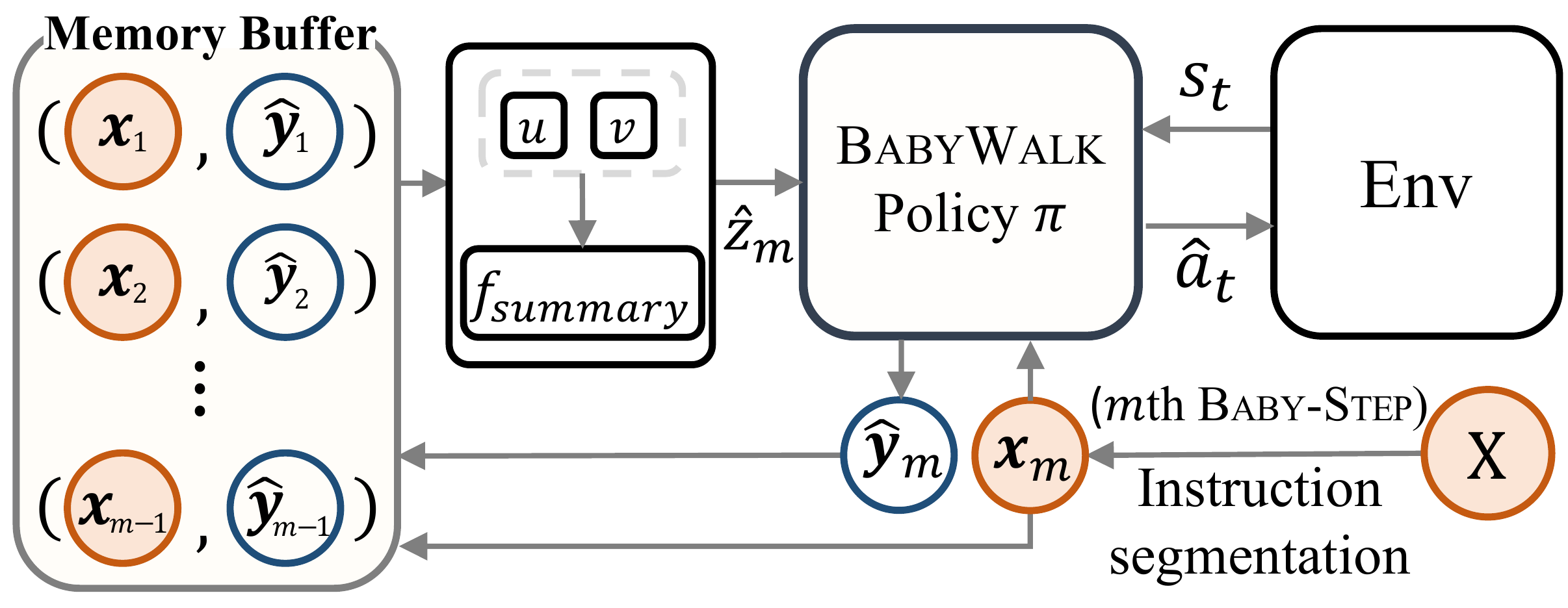}

    \caption{
    \textbf{The \bbwalk agent} has a memory buffer storing its past experiences of instructions $\bsx_m$, and its trajectory $\hat{\bsy}_m$. When a new \bbstep $\bsx_m$ is presented, the agent retrieves from the memory a summary of its experiences as the history context. It takes actions conditioning on the context (as well as its state $\bss_t$ and the previous action $\hat{\bsa}_t$). Upon finishing following the instruction. the trajectory $\hat{\bsy}_m$ is then sent to the memory to be remembered.}
    \label{fig:agent}
  
\end{figure}

\subsection{The \bbwalk Agent}
\label{subsec:agent}

The basic operating model of our navigation agent \bbwalk is to follow a ``micro instruction'' $\bsx_m$ (\ie, a short sequence of instructions, to which we also refer as \bbstep), conditioning on the context $\hat{\bsz}_m$ and to output a trajectory $\hat{\bsy}_m$. A schematic diagram is shown in Fig.~\ref{fig:agent}. Of particularly different from previous approaches is the introduction of a novel memory module. We assume the \bbsteps are given in the training and inference time -- \S~\ref{sec:dataset} explains how to obtain them if not given \emph{a prior} (Readers can directly move to that section and return to this part afterwards). The left of the Fig.~\ref{fig:curriculum} gives an example of those micro-instructions.

\mypara{Context} The context is a summary of the past experiences of the agent, namely the previous $(m-1)$ mini-instructions and  trajectories:
\begin{align}
    \hat{\bsz}_m = {g}\big( & \summary(\bsx_1,\cdots, \bsx_{m-1}),\nonumber \\
    & \summary(\hat{\bsy}_1,\cdots, \hat{\bsy}_{m-1}) \big)
\label{eq:context}
\end{align}
where the function $g$ is implemented with a multi-layer perceptron. The summary function $\summary$ is explained in below.

\mypara{Summary} To map variable-length sequences (such as the trajectory and the instructions) to a single vector, we can use various mechanisms such as LSTM. We reported an ablation study on this in \S~\ref{sec:analysis}. In the following, we describe the ``forgetting'' one that weighs more heavily towards the most recent experiences and performs the best empirically.
\begin{align}
    \summary(\bsx_1,\cdots, \bsx_{m-1}) &= \sum_{i=1}^{m-1} \alpha_i \cdot u(\bsx_i)  \label{eq:memory_x} \\ 
    \summary(\hat{\bsy}_1,\cdots, \hat{\bsy}_{m-1}) &= \sum_{i=1}^{m-1} \alpha_i \cdot v(\hat{\bsy}_i)
    \label{eq:memory_y}
\end{align}
where the weights are normalized to 1 and inverse proportional to how far $i$ is from $m$,
\begin{equation}
\alpha_i \propto \exp{\big(-\gamma \cdot \omega(m-1-i)\big)}
\end{equation}
$\gamma$ is a hyper-parameter (we set to $1/2$) and $\omega(\cdot)$ is a monotonically nondecreasing function and we simply choose the identity function. 

Note that, we summarize over representations of ``micro-instructions'' ($\bsx_m$) and experiences of executing those micro-instructions $\hat{\bsy}_m$. The two encoders $u(\cdot)$ and $v(\cdot)$ are described in \S~\ref{subsec:impl}. They are essentially the summaries of ``low-level'' details, \ie, representations of a sequence of words, or a sequence of states and actions.  While existing work often directly summarizes all the low-level details, we have found that  the current form of ``hierarchical'' summarizing (\ie, first summarizing each \bbstep, then summarizing all previous \bbsteps) performs better.   

\mypara{Policy} The agent takes actions, conditioning on the context $\hat{\bsz}_m$, and the current instruction $\bsx_m$:
\begin{align}
	 \hat{\bsa}_t & \sim \bspi\paren{\cdot | \bss_{t}, \hat{\bsa}_{t-1};   u(\bsx_{m}), \hat{\bsz}_{m}} 
	\label{eqn:agent}
\end{align}
where the policy is implemented with a LSTM with the same cross-modal attention between visual states and languages as in~\citep{fried2018speaker}. 

%% file: method_algo.tex
% !TEX root = main.tex
\subsection{Learning of the \bbwalk Agent}
\label{subsec:algo}
\begin{figure*}[t]
    \centering
    {
        \includegraphics[width=\textwidth]{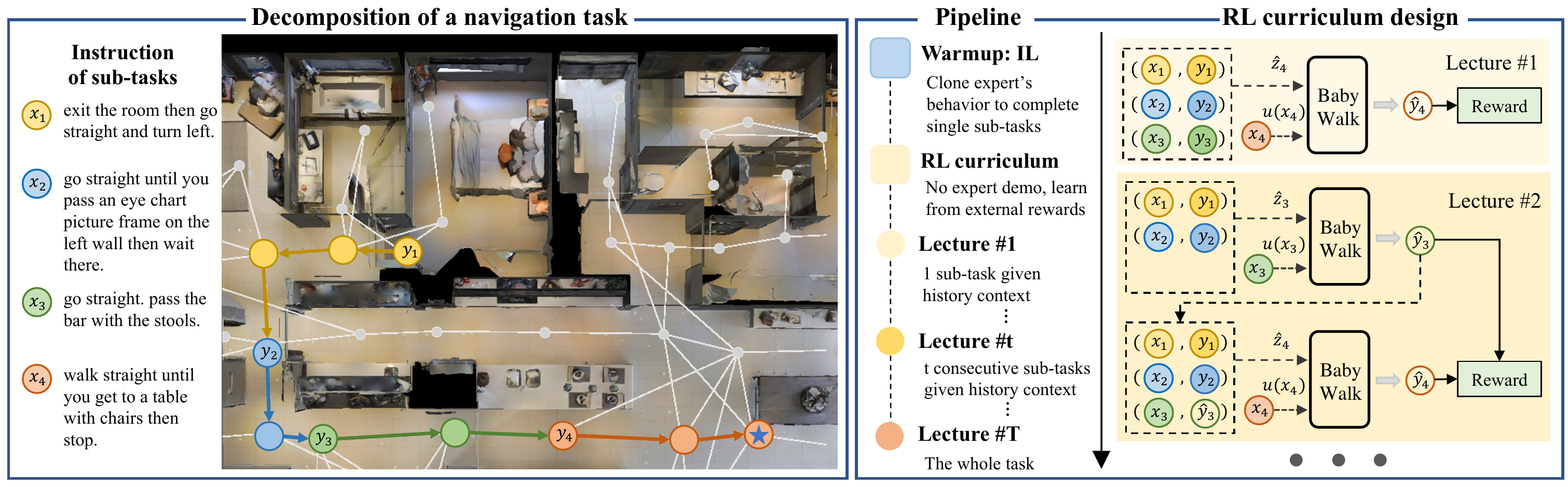}
    }

    \caption{
    \textbf{Two-phase learning by \bbwalk.} (Left) An example instruction-trajectory pair from the \rfourr dataset is shown. The long instruction is segmented into four \bbstep instructions. We use those \bbsteps for imitation learning (\S~\ref{subsec:imitation})  (Right) Curriculum-based RL. The \bbwalk agent warm-starts from the imitation learning policy, and incrementally learns to handle longer tasks by executing consecutive \bbsteps and getting feedback from external rewards (\cf \S~\ref{subsec:reinforce}). We illustrate two initial RL lectures using the left example.}
    \label{fig:curriculum}

\end{figure*}

The agent learns in  two phases. In the first one, imitation learning is used where the agent learns to execute \bbsteps accurately. In the second one, the agent learns to execute successively longer tasks from a designed curriculum.

\subsubsection{Imitation Learning}
\label{subsec:imitation}

\bbsteps are shorter navigation tasks. With the $m$th instruction $\bsx_m$, the agent is asked to follow the instruction so that its trajectory matches the human expert's $\bsy_m$. To assist the learning, the context is computed from the human expert trajectory up to the $m$th \bbstep (\ie, in eq.~(\ref{eq:context}), $\hat{\bsy}$s are replaced with $\bsy$s). We maximize the objective  
\begin{equation}
 \ell = \sum_{m=1}^\rmM \sum_{t_m=1}^{\abs{\bsy_m}} \log \bspi\paren{ \bsa_{t_m} | \bss_{t_m}, \bsa_{t_m - 1} ;  u(\bsx_{m}), \bsz_{m}} \nonumber
\end{equation}
We emphasize here each \bbstep is treated independently of the others in this learning regime. Each time a \bbstep is to be executed, we ``preset'' the agent in the human expert's context and the last visited state. We follow existing literature~\citep{anderson2018vision,fried2018speaker} and use student-forcing based imitation learning, which uses agent's predicted action instead of the expert action for the trajectory rollout. 

\subsubsection{Curriculum Reinforcement Learning}
\label{subsec:reinforce}

We want the agent to be able to execute multiple consecutive \bbsteps and optimize its performance on following longer navigation instructions (instead of the cross-entropy losses from the imitation learning).  However, there is a  discrepancy between our goal of training the agent to cope with the uncertainty in a long instruction and the imitation learning agent's ability in accomplishing shorter tasks \emph{given the human annotated history}. Thus it is challenging to directly optimize the agent with a typical RL learning procedure, even the imitation learning might have provided a good initialization for the policy, see our ablation study in \S~\ref{sec:analysis}.

Inspired by the curriculum learning strategy~\citep{bengio2009curriculum}, we design an incremental learning process that the agent is presented with a curriculum of increasingly longer navigation tasks. Fig.~\ref{fig:curriculum} illustrates this idea with two ``lectures''.
Given a long navigation instruction $\rmX$ with $\rmM$ \bbsteps, for the $k$th lecture, the agent is given all the human expert's trajectory up to but not including the $(\rmM-k+1)$th \bbstep, as well as the history context $\bsz_{\rmM-k+1}$. The agent is then asked to execute the $k$th micro-instructions from $\bsx_{\rmM-k+1}$ to $\bsx_{\rmM}$ using reinforcement learning to produce its trajectory that optimizes a task related metric, for instance the fidelity metric measuring how faithful the agent follows the instructions.

As we increase $k$ from 1 to $\rmM$, the agent faces the challenge of navigating longer and longer tasks with reinforcement learning. However, the agent only needs to improve its skills from its prior exposure to shorter ones.  Our ablation studies show this is indeed a highly effective strategy.

%% file: method_dataset.tex
% !TEX root = main.tex

\subsection{New Datasets for Evaluation \&  Learning}
\label{sec:dataset}

\begin{table}[t]
\centering
\small
\tabcolsep 5pt
\resizebox{1.0\linewidth}{!}
{
\begin{tabular}{@{\;}l@{\;\;}cccc@{\;}}
\toprule
& \rtwor & \rfourr & \rsixr & \reightr \\
\midrule
Train seen instr. & 14,039 & 233,532 & 89,632 & 94,731 \\
Val unseen instr. & 2,349 & 45,234 & 35,777 & 43,273\\
Avg instr. length & 29.4 & 58.4 & 91.2 & 121.6 \\
\midrule
Avg \# \bbsteps & 1.8 & 3.6 & 5.6 & 7.4 \\
\bottomrule
\end{tabular}
}
\caption{Datasets used for VLN learning and evaluation}
\label{tbl:datasets}
\end{table}

\begin{figure}[t]
    \includegraphics[width=0.475\textwidth]{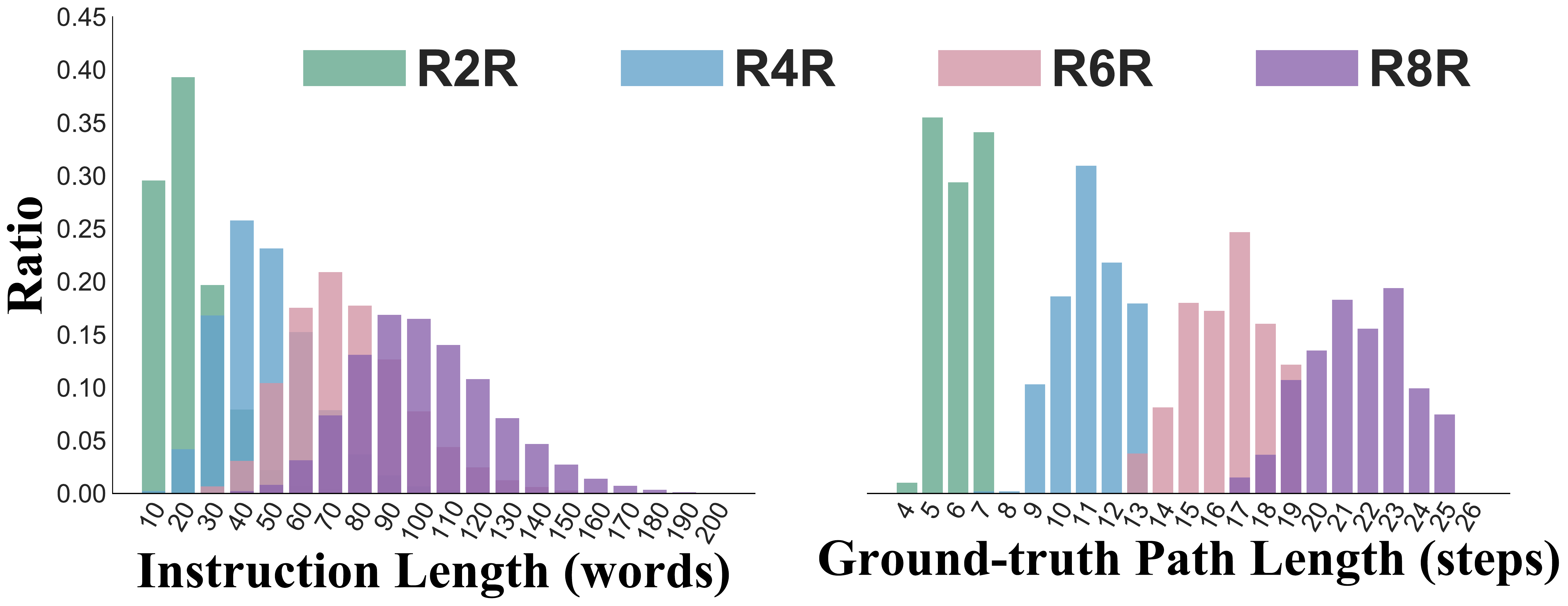}
    \caption{
    \small The distribution of lengths of instructions  and ground-truth trajectories in our datasets.}
    \label{fig:distribution}
\end{figure}

To our best knowledge, this is the first work studying how well VLN agents generalize to long navigation tasks. To this end, we create the following datasets in the same style as in~\citep{jain2019stay}. 

\mypara{\RsixR and \ReightR} We concatenate the trajectories in the training as well as the validation unseen split of the \RtwoR dataset for 3 times and 4 times respectively, thus extending the lengths of navigation tasks to 6 rooms and 8 rooms. To join, the end of the former trajectory must be within 0.5 meter with the beginning of the later trajectory. Table~\ref{tbl:datasets} and Fig.~\ref{fig:distribution} contrast the different datasets in the \# of instructions, the average length (in words) of instructions and how the distributions vary. 

Table~\ref{tbl:datasets} summarizes the descriptive statistics of \bbsteps across all datasets used in this paper. The datasets and the segmentation/alignments are made publically available\footnote{Available at \url{https://github.com/Sha-Lab/babywalk}}. 

%% file: method_impl.tex
% !TEX root = main.tex

\subsection{Key Implementation Details}
\label{subsec:impl}

In the following, we describe key information for research reproducibility, while the complete details are in the \SM. 

\mypara{States and Actions} We follow~\cite{fried2018speaker} to set up the states as the visual features (\ie ResNet-152 features~\cite{he2016deep}) from the agent-centric panoramic  views in 12 headings $\times$ 3 elevations with 30 degree intervals. Likewise, we use the same panoramic action space.

\mypara{Identifying \bbsteps} Our learning approach requires an agent to follow micro-instructions (\ie, the \bbsteps). Existing datasets~\citep{anderson2018vision,jain2019stay,chen2019touchdown} do not provide fine-grained segmentations of long instructions. Therefore, we use a template matching approach to aggregate consecutive sentences into \bbsteps. First, we extract the noun phrase using POS tagging. Then, we employs heuristic rules to chunk a long instruction into shorter segments according to punctuation and landmark phrase (\ie, words for concrete objects). We document the details in the \SM. 

\mypara{Aligning \bbsteps with Expert Trajectory} \textit{Without extra annotation}, we propose a method to approximately chunk original expert trajectories into sub-trajectories that align with the \bbsteps. This is important for imitation learning at the micro-instruction level (\S~\ref{subsec:imitation}). Specifically, we learn a multi-label visual landmark classifier to identify concrete objects from the states along expert trajectories by using the landmark phrases extracted from the their instructions as weak supervision. For each trajectory-instruction pair, we then extract the visual landmarks of every state as well as the landmark phrases in \bbstep instructions. Next, we perform a  dynamic programming procedure to segment the expert trajectories by aligning the visual landmarks and landmark phrases, using the confidence scores of the multi-label visual landmark classifier to form the function.

\mypara{Encoders and Embeddings} The encoder $u(\cdot)$ for the (micro)instructions is a LSTM. The encoder for the trajectory $\bsy$ contains two separate Bi-LSTMs, one for the state $\bss_t$ and the other for the action $\bsa_t$. The outputs of the two Bi-LSTMs are then concatenated to form the embedding function $v(\cdot)$. The details of the neural network architectures (\ie configurations as well as an illustrative figure), optimization hyper-parameters, \etc are included in the \SM. 

\mypara{Learning Policy with Reinforcement Learning} In the second phase of learning, \bbwalk uses RL to learn a policy that maximizes the \emph{fidelity-oriented} rewards (CLS) proposed by~\citet{jain2019stay}. We use policy gradient as the optimizer~\citep{sutton2000policy}. Meanwhile, we set the maximum number of lectures in curriculum RL to be 4, which is studied in Section~\ref{sec:analysis}.

%% file: exp.tex
% !TEX root = main.tex

\input{exp_big_table}

\section{Experiments}
\label{sec:experiment}

We describe the experimental setup (\S~\ref{sec:setups}),followed by the main results in \S~\ref{sec:results} where we show the proposed \bbwalk agent attains competitive results on both the in-domain dataset but also generalizing to out-of-the-domain datasets with varying lengths of navigation tasks. We report results from various ablation studies in \S~\ref{sec:analysis}. While we primarily focus on the \RfourR dataset, we re-analyze the original \RtwoR dataset in \S~\ref{sec:r2r} and were surprised to find out the agents trained on it can generalize.

\input{exp_setups}

\input{exp_results}

\input{exp_analysis}

\input{exp_r2r}

%% file: exp_big_table.tex
% !TEX root = main.tex

\begin{table*}[ht]
    \centering
	\tabcolsep 2pt
    \resizebox{1.0\linewidth}{!}
    {
    \small
	    \begin{tabular}{@{\;}l@{\;\;} ccc @{\quad} ccc @{\quad} ccc @{\quad} ccc @{\quad} ccc @{\;}}
	        \addlinespace
	        \toprule
	        & \multicolumn{3}{c}{{\bf In-domain}} & \multicolumn{12}{c}{{\bf Generalization to other datasets}} \\
	        Setting & \multicolumn{3}{c}{\rfourr $\rightarrow$ \rfourr} & \multicolumn{3}{c}{\rfourr $\rightarrow$ \rtwor} & \multicolumn{3}{c}{\rfourr $\rightarrow$ \rsixr} & \multicolumn{3}{c}{\rfourr $\rightarrow$ \reightr} & \multicolumn{3}{c}{Average} \\\cmidrule(lr){2-4}\cmidrule(lr){5-16}
	        Metrics & \textsc{sr}$\uparrow$ & \textsc{cls}$\uparrow$ & \textsc{sdtw}$\uparrow$ & \textsc{sr}$\uparrow$ & \textsc{cls}$\uparrow$ & \textsc{sdtw}$\uparrow$ &
	        \textsc{sr}$\uparrow$ & \textsc{cls}$\uparrow$ & \textsc{sdtw}$\uparrow$ &
	        \textsc{sr}$\uparrow$ & \textsc{cls}$\uparrow$ & \textsc{sdtw}$\uparrow$ &
			\textsc{sr}$\uparrow$ & \textsc{cls}$\uparrow$ & \textsc{sdtw}$\uparrow$ \\
	        \midrule
			\text{\textsc{seq2seq}} & 25.7 & 20.7 & 9.0 & 16.3 & 27.1 & 10.6 & 14.4 & 17.7 & 4.6 & 20.7 & 15.0 & 4.7 & 17.1 & 19.9 & 6.6 \\
			\text{\textsc{sf}$^+$} & 24.9 & 23.6 & 9.2 & 22.5 & 29.5 & 14.8 & 15.5 & 20.4 & 5.2 & 21.6 & 17.2 & 5.0 & 19.9 & 22.4 & 8.3 \\
			\text{\textsc{rcm(goal)}$^+$} & 28.7 & 36.3 & 13.2 & 25.9 & 44.2 & 20.2 & 19.3 & 31.8 & 7.3 & 22.8 & 27.6 & 5.1 & 22.7 & 34.5 & 10.9 \\
			\text{\textsc{rcm(fidelity)}$^+$} & 24.7 & 39.2 & 13.7 & 29.1 & 34.3 & 18.3 & 20.5 & 38.3 & 7.9 & 20.9 & 34.6 & 6.1 & 23.5 & 35.7 & 10.8 \\
			\text{\textsc{regretful}$^{+\star}$} & 30.1 & 34.1 & 13.5 & 22.8 & 32.6 & 13.4 & 18.0 & 31.7 & 7.5 & 18.7 & 29.3 & 5.6 & 19.8 & 31.2 & 8.8 \\
			\text{\textsc{fast}$^{+\star}$} & \bf 36.2 & 34.0 & 15.5 & 25.1 & 33.9 & 14.2 & 22.1 & 31.5 & 7.7 & \bf 27.7 & 29.6 & 6.3 & 25.0 & 31.7 & 9.4 \\
	        \midrule
			 {\ourmethod} & 29.6 & 47.8 & \bf 18.1 & \bf 35.2 & 48.5 & 27.2 & \bf 26.4 & 44.9 & 13.1 & 26.3 & 44.7 & \bf 11.5 & \bf 29.3 & 46.0 & 17.3 \\
			 {\ourmethod$^+$} & 27.3 & \bf 49.4 & 17.3 & 34.1 & \bf 50.4 & \bf 27.8 & 25.5 & \bf 47.2 & \bf 13.6 & 23.1 & \bf 46.0 & 11.1 & 27.6 & \bf 47.9 & \bf 17.5 \\
	     \bottomrule
	    \end{tabular}
	}
	\caption{
    	VLN agents trained on the \rfourr dataset and evaluated on the unseen portion of the \rfourr (in-domain) and the other 3 out-of-the-domain datasets: \rtwor, \rsixr and \reightr with different distributions in instruction length. The \SM has more comparisons. ($^+$: pre-trained with data augmentation. $^\star$: reimplemented or adapted from the original authors' public codes).  
    }
   
    \label{tab:transfer_r4r}
\end{table*}

%% file: exp_setups.tex
% !TEX root = main.tex
\subsection{Experimental Setups.}
\label{sec:setups}

\mypara{Datasets} We conduct empirical studies on the existing datasets \RtwoR and \RfourR\citep{anderson2018vision,jain2019stay}, and the two newly created benchmark datasets \RsixR and \ReightR, described in ~\S~\ref{sec:dataset}. Table~\ref{tbl:datasets} and Fig. ~\ref{fig:distribution} contrast their differences.

\mypara{Evaluation Metrics} We adopt the following metrics: \emph{Success Rate (\textsc{sr})} that measures the average rate of the agent stopping within a specified distance near the goal location~\citep{anderson2018vision}, \emph{Coverage weighted by Length Score (\textsc{cls})}~\citep{jain2019stay} that measures the fidelity of the agent's path to the reference, weighted by the length score, and the newly proposed \emph{Success rate weighted normalized Dynamic Time Warping (\textsc{sdtw})} that measures in more fine-grained details, the spatio-temporal similarity of the paths by the agent and the human expert, weighted by the success rate~\citep{magalhaes2019effective}. Both \textsc{cls} and \textsc{sdtw} measure explicitly the agent's ability to follow instructions and in particular, it was shown that \textsc{sdtw} corresponds to human preferences the most. We report results in other metrics in the \SM.

\mypara{Agents to Compare to} Whenever possible, for all agents we compare to, we either re-run, reimplement or adapt publicly available codes from their corresponding authors with their provided instructions to ensure a fair comparison. We also ``sanity check'' by ensuring the results from our implementation and adaptation replicate and are comparable to  the  reported ones in the literature.

We compare our \bbwalk to the following:  (1) the \textsc{seq2seq} agent~\citep{anderson2018vision}, being adapted to the panoramic state and action space used in this work; (2) the Speaker Follower (\textsc{sf}) agent~\citep{fried2018speaker}; (3) the Reinforced Cross-Modal Agent (\textsc{rcm})~\citep{wang2019reinforced} that refines the \textsc{sf} agent using reinforcement learning with either \emph{goal-oriented reward} (\textsc{rcm(goal)}) or \emph{fidelity-oriented reward} (\textsc{rcm(fidelity)}); (4)
the Regretful Agent (\textsc{regretful})~\citep{ma2019regretful} that uses a progress monitor that records visited path and a regret module that performs backtracking; (5) the Frontier Aware Search with Backtracking agent (\textsc{fast})~\citep{ke2019tactical} that incorporates global and local knowledge to compare partial trajectories in different lengths.

The last 3 agents are reported having state-of-the art results on the benchmark datasets.  Except the \textsc{seq2seq} agent, all other agents depend on an additional pre-training stage with data augmentation~\citep{fried2018speaker}, which improves cross-board. Thus, we train two \bbwalk agents: one with and the other without the data augmentation.

%% file: exp_results.tex
% !TEX root = main.tex
\subsection{Main results}
\label{sec:results}

\mypara{In-domain Generalization} This is the standard evaluation scenario where a trained agent is assessed on the unseen split from the same dataset as the training data. The leftmost columns in Table~\ref{tab:transfer_r4r} reports the results where the training data is from \rfourr.  The \bbwalk agents outperform  all other agents when evaluated on \textsc{cls} and \textsc{sdtw}.

When evaluated on \textsc{sr},  \textsc{fast} performs the best and the \bbwalk agents do not stand out. This is expected: agents which are trained to reach goal do not necessarily lead to better instruction-following. Note that \textsc{rcm(fidelity)} performs well in path-following.

\begin{figure}[t]
    \centering
    \tabcolsep 1pt
    \includegraphics[width=0.475\textwidth]{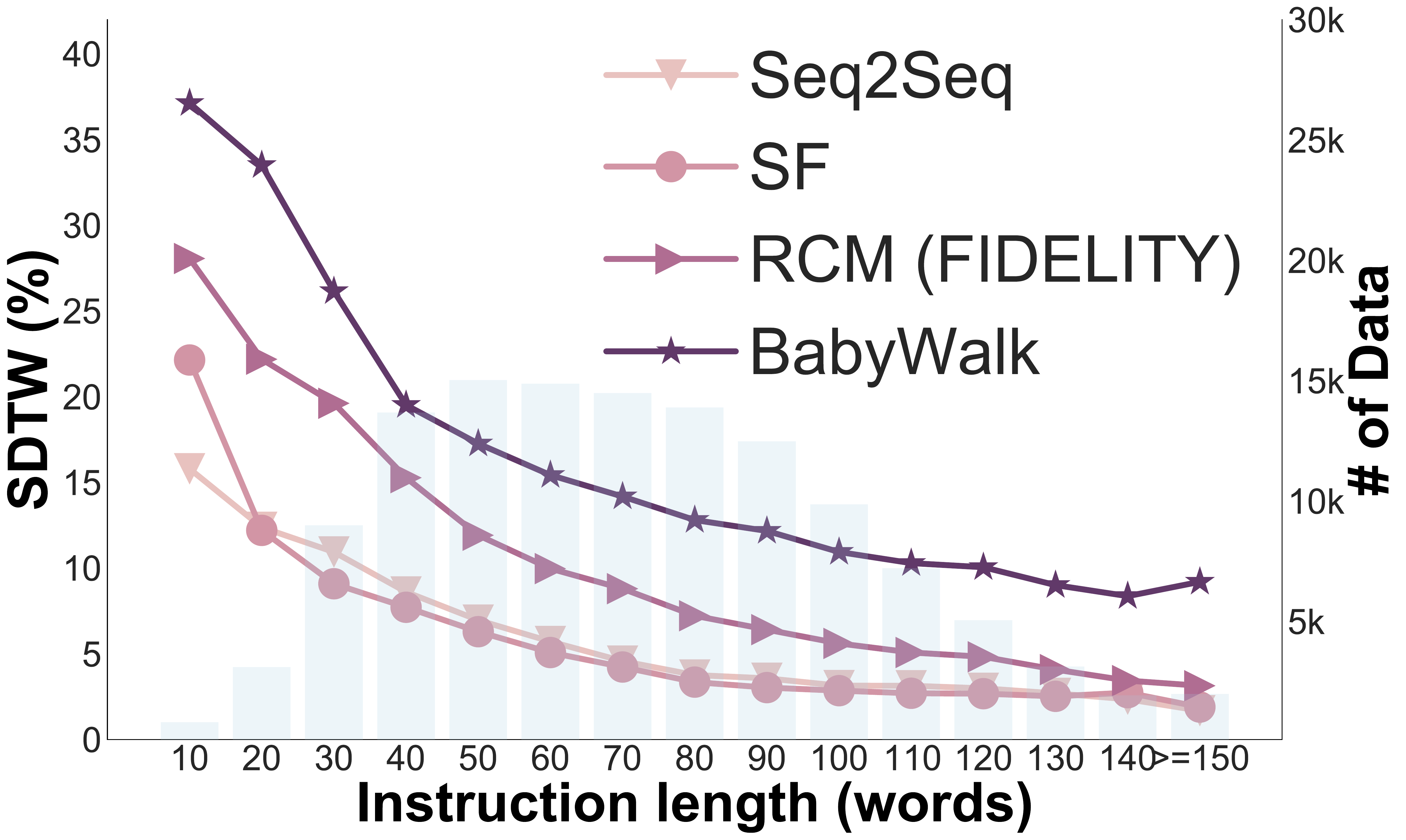}

    \caption{
    Performance by various agents on navigation tasks in different lengths.  See texts for details. 
    }
    \label{fig:cross-horizon}
\end{figure}

\mypara{Out-of-domain Generalization} While our primary goal is to train agents to generalize well to longer navigation tasks, we are also curious how the agents perform on shorter navigation tasks too. The right columns in Table~\ref{tab:transfer_r4r} report the comparison. The \bbwalk agents outperform all other agents in all  metrics except \textsc{sr}. In particular, on \textsc{sdtw}, the generalization to \rsixr and \reightr is especially encouraging, resulting almost twice those of the second-best agent \textsc{fast}. Moreover, recalling from Fig.~\ref{fig:teaser}, \bbwalk's generalization to \rsixr and \reightr attain even better performance than the \textsc{rcm} agents that are trained \emph{in-domain}.

Fig.~\ref{fig:cross-horizon} provides additional evidence on the success of \bbwalk, where we have contrasted to its performance to other agents' on following instructions in different lengths \emph{across all datasets}. Clearly, the \bbwalk agent is able to improve very noticeably on longer instructions.   
 
\mypara{Qualitative Results} Fig.~\ref{fig:qualitativesmall} contrasts visually several agents in executing two (long) navigation tasks. \bbwalk's trajectories are similar to what human experts provide, while other agents' are not.

\def\VizImageWidth{1.3in}
\def\VizImageHeightLong{1.3in}

\begin{figure*}[t]
\small
\centering
\tabcolsep 1pt

\resizebox{0.98\linewidth}{!}{ \small

\begin{tabular}{@{\;}ccccc}
    \toprule
    \footnotesize{\textsc{human}} & \bbwalk & \footnotesize{\textsc{rcm}} & \footnotesize{\textsc{sf}} & \footnotesize{\textsc{seq2seq}} \\ \midrule
    \includegraphics[width=\VizImageWidth,height=\VizImageHeightLong]{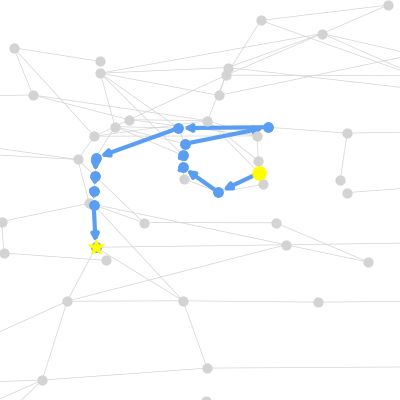} &
    \includegraphics[width=\VizImageWidth,height=\VizImageHeightLong]{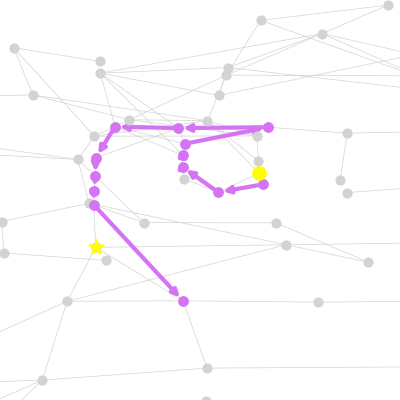} &
    \includegraphics[width=\VizImageWidth,height=\VizImageHeightLong]{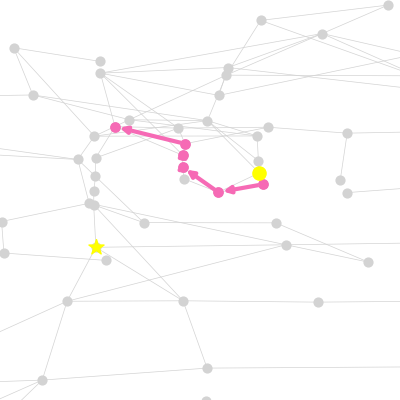} &
    \includegraphics[width=\VizImageWidth,height=\VizImageHeightLong]{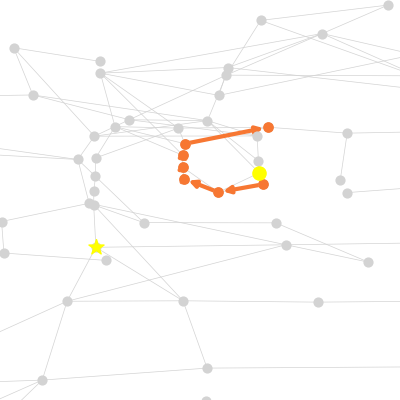} &
    \includegraphics[width=\VizImageWidth,height=\VizImageHeightLong]{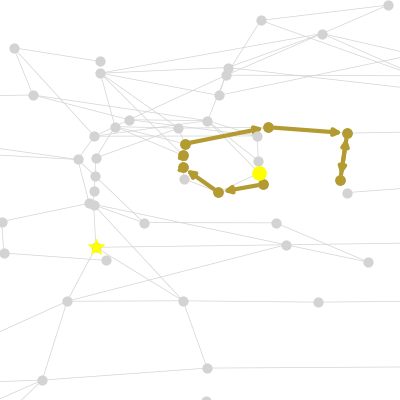} \\

    \includegraphics[width=\VizImageWidth,height=\VizImageHeightLong]{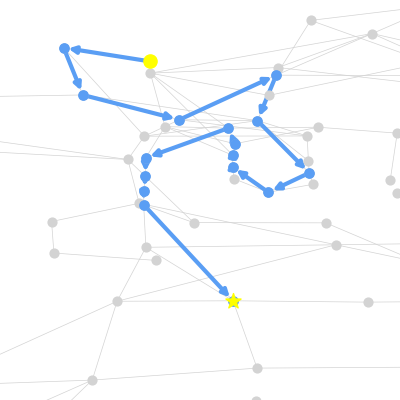} &
    \includegraphics[width=\VizImageWidth,height=\VizImageHeightLong]{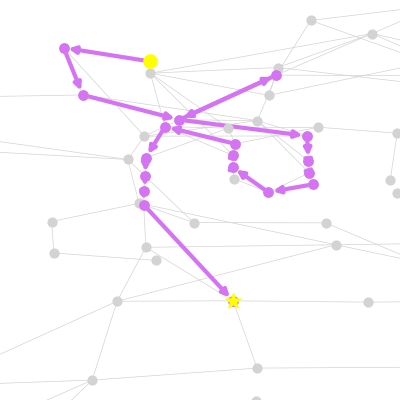} &
    \includegraphics[width=\VizImageWidth,height=\VizImageHeightLong]{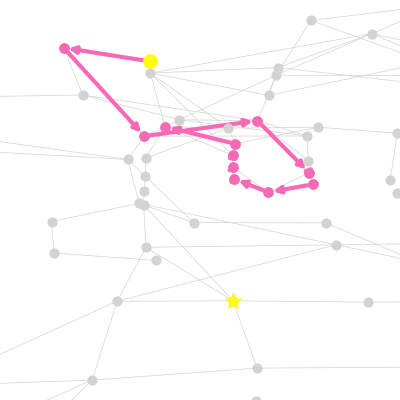} &
    \includegraphics[width=\VizImageWidth,height=\VizImageHeightLong]{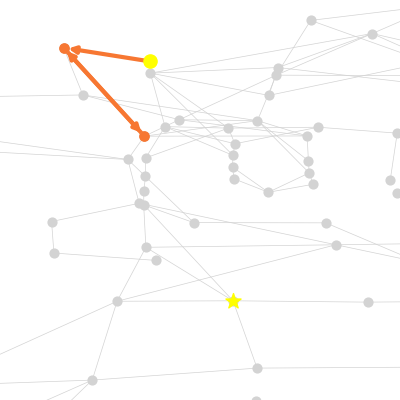} &
    \includegraphics[width=\VizImageWidth,height=\VizImageHeightLong]{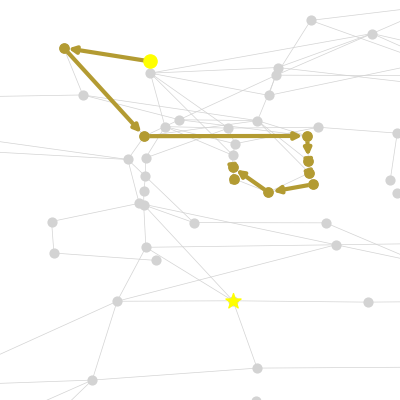} \\
    \bottomrule
\end{tabular}
}
\caption{
    Trajectories by human experts and VLN agents on two navigation tasks. More are in the \SM.
}

\label{fig:qualitativesmall}
\end{figure*}

%% file: exp_analysis.tex
% !TEX root = main.tex
\subsection{Analysis}
\label{sec:analysis}
\begin{table}[t]
    \centering
    \small
	\tabcolsep 3pt
    \resizebox{1.0\linewidth}{!}
    {
    \begin{tabular}{@{\;}l@{\quad}ccc @{\quad} ccc@{\;}}
        \addlinespace
        \toprule
        Setting & \multicolumn{3}{c}{\rfourr $\rightarrow$ \rfourr} & \multicolumn{3}{c}{\rfourr $\rightarrow$ others} \\
        Metrics & \textsc{sr}$\uparrow$ & \textsc{cls}$\uparrow$ & \textsc{sdtw} $\uparrow$ & \textsc{sr}$\uparrow$ & \textsc{cls}$\uparrow$ & \textsc{sdtw} $\uparrow$ \\
        \midrule
		\multicolumn{7}{@{}l}{$\summary = $} \\
        \textsc{null} & 18.9 & 43.1 & 9.9 & 17.1 & 42.3 & 9.6 \\
        $\textsc{lstm}(\cdot)$ & 25.8 & 44.0 & 14.4 & 25.7 & 42.1 & 14.3 \\
        \midrule
        \multicolumn{7}{@{}l}{$\summary = \sum_{i=1}^{m-1} \alpha_i \cdot (\cdot)$, \ie, eqs.~(\ref{eq:memory_x},\ref{eq:memory_y})} \\[4pt]
        \quad $\gamma = 5$ & 27.5 & 46.8 & 15.8 & 26.7 & 44.4 & 14.9 \\
        \quad $\gamma = 0.5$  & 27.3 & \bf 49.4 & \bf 17.3 & \bf 27.6 & \bf 47.9 & \bf 17.5 \\
        \quad $\gamma = 0.05$ & \bf 27.5 & 47.7 & 16.2 & 26.0 & 45.5 & 15.2 \\
        \quad $\gamma = 0$ & 26.1 & 46.6 & 15.1 & 25.1 & 44.3 & 14.4 \\
     \bottomrule
    \end{tabular}
    }
    \caption{
    	The memory buffer is beneficial to generalizing to different tasks from on which the agent is trained.
    }
    \label{tab:context}
   
\end{table}

\mypara{Memory Buffer is Beneficial} Table~\ref{tab:context} illustrates the importance of having a memory buffer to summarize the agent's past experiences. Without the memory (\textsc{null}), generalization to longer tasks is significantly worse. Using LSTM to summarize is worse than using forgetting to summarize (eqs.~(\ref{eq:memory_x},\ref{eq:memory_y})). Meanwhile, ablating $\gamma$ of the forgetting mechanism concludes that $\gamma = 0.5$ is the optimal to our hyperparameter search. Note that when $\gamma = 0$, this mechanism degenerates to taking average of the memory buffer, and leads to inferior results.

\mypara{Curriculum-based RL (CRL) is Important} Table~\ref{tab:curriculum} establishes the value of CRL. While imitation learning (\textsc{il}) provides a good warm-up for \textsc{sr}, significant improvement on other two metrics come from the subsequent RL (\textsc{il+rl}). Furthermore, CRL (with 4 ``lectures'') provides clear improvements over direct RL on the entire instruction (\ie, learning to execute all \bbsteps at once). Each lecture improves over the previous one, especially in terms of the \textsc{sdtw} metric.

\begin{table}[t]
    \centering
    \small
	\tabcolsep 5pt
    \small
    \resizebox{1.0\linewidth}{!}
    {
    \begin{tabular}{@{\;}l@{\quad}ccc @{\quad} ccc@{\;}}
        \addlinespace
        \toprule
        Setting & \multicolumn{3}{c}{\rfourr $\rightarrow$ \rfourr} & \multicolumn{3}{c}{\rfourr $\rightarrow$ others} \\
        Metrics & \textsc{sr}$\uparrow$ & \textsc{cls}$\uparrow$ & \textsc{sdtw} $\uparrow$ & \textsc{sr}$\uparrow$ & \textsc{cls}$\uparrow$ & \textsc{sdtw} $\uparrow$ \\
        \midrule 
        		\textsc{il}	 & 24.7 & 27.9 & 11.1 & 24.2 & 25.8 & 10.2 \\
		\textsc{il}+\textsc{rl}  & 25.0 & 45.5 & 13.6 & 25.0 & 43.8 & 14.1 \\
		\midrule
		\multicolumn{7}{@{}l}{\textsc{il}+ \textsc{crl} w/ \textsc{lecture }\#} \\

		1\textit{st} 		   & 24.1 & 44.8 & 13.5 & 24.1 & 43.1 & 13.6 \\
		2\textit{nd} 		 & 26.7 & 45.9 & 15.2 & 26.2 & 43.7 & 14.8 \\
		3\textit{rd} 		 & \bf 27.9 & 47.4 & 17.0 & 26.7 & 45.4 & 16.3 \\
		4\textit{th}  	     & 27.3 & \bf 49.4 & \bf 17.3 & \bf 27.6 & \bf 47.9 & \bf 17.5 \\
     \bottomrule
    \end{tabular}
    }
    \caption{
        \bbwalk's performances with curriculum-based reinforcement learning (\textsc{crl}), which improves imitation learning without or with  reinforcement learning  (\textsc{il+rl}).  
    }
  
    \label{tab:curriculum}
   \end{table}

%% file: exp_r2r.tex
% !TEX root = main.tex
\subsection{Revisiting \RtwoR}
\label{sec:r2r}
\begin{table}[t]
	\centering
	\small
	\tabcolsep 5pt
	
	\resizebox{1.0\linewidth}{!}
	{
	\begin{tabular}{@{\;}l@{\;\;}ccc @{\quad} ccc @{\;}}
			\addlinespace
			\toprule
			Eval &   \multicolumn{3}{c}{$\rightarrow$ \rsixr} & \multicolumn{3}{c}{ $\rightarrow$ \reightr} \\
			Training  & \textsc{sr}$\uparrow$ & \textsc{cls}$\uparrow$ & \textsc{sdtw}$\uparrow$ &
			\textsc{sr}$\uparrow$ & \textsc{cls}$\uparrow$ & \textsc{sdtw}$\uparrow$  \\
			\midrule
			\rtwor & 21.7 & \bf 49.0 & 11.2 & 20.7 & \bf 48.7 & 9.8 \\
			\rfourr & \bf 25.5 & 47.2 & \bf 13.6 & \bf 23.1 & 46.0 & \bf 11.1\\
			\bottomrule
			\addlinespace
			
			Eval &   \multicolumn{3}{c}{$\rightarrow$ \rtwor} & \multicolumn{3}{c}{ $\rightarrow$ \rfourr} \\
			Training  & \textsc{sr}$\uparrow$ & \textsc{cls}$\uparrow$ & \textsc{sdtw}$\uparrow$ & \textsc{sr}$\uparrow$
			& \textsc{cls}$\uparrow$ & \textsc{sdtw}$\uparrow$  \\
			
			\midrule
			\rtwor & \bf 43.8 & \bf 54.4 & \bf 36.9 & 21.4 & \bf 51.0 & 13.8  \\
			\rfourr & 34.1 & 50.4 & 27.8 & \bf 27.3 & 49.4 & \bf 17.3 \\
			\bottomrule

		\end{tabular}
	}
	\caption{
		(Top) \bbwalk trained on \rtwor is nearly as effective as the agent trained on \rfourr when generalizing to longer tasks. (Bottom) \bbwalk  trained on \rtwor adapts to \rfourr better than the agent trained in the reverse direction. 
	}

	\label{tab:r2r_other}
\end{table}

Our experimental study has been focusing on using \rfourr as the training dataset as it was established that as opposed to \rtwor, \rfourr distinguishes well an agent who just learns to reach the goal from an agent who learns to follow instructions.

Given the encouraging results of generalizing to longer tasks, a natural question to ask, \emph{how well can an agent trained on \rtwor generalize?}

Results in Table~\ref{tab:r2r_other} are interesting. Shown in the top panel, the difference in the averaged performance of generalizing to \rsixr and \reightr is not significant. The agent trained on \rfourr has a small win on \rsixr presumably because \rfourr is closer to \rsixr than \rtwor does. But for even longer tasks in \reightr, the win is similar. 

In the bottom panel,  however, it seems that  \rtwor $\rightarrow$ \rfourr  is stronger (incurring less loss in performance when compared to the in-domain setting \rfourr $\rightarrow$ \rfourr) than the reverse direction (\ie, comparing \rfourr $\rightarrow$ \rtwor to the in-domain \rtwor $\rightarrow$ \rtwor). This might have been caused by the noisier segmentation of long instructions into \bbsteps in \rfourr. (While \rfourr is composed of two navigation paths in \rtwor, the segmentation algorithm is not aware of the ``natural'' boundaries between the two paths.)

%% file: discussion.tex
% !TEX root = main.tex
\section{Discussion}
\label{sec:discussion}

There are a few future directions to pursue. First, despite the significant improvement, the  gap between short and long tasks is still large and needs to be further reduced. 
Secondly, richer and more complicated variations between the learning setting and the real  physical world need to be tackled. For instance, developing agents that are robust to variations in both visual appearance and instruction descriptions is an important next step.

%% file: supp_content.tex
% !TEX root = supp.tex

\noindent In this supplementary material, we provide details omitted in the main text. The content is organized as what follows:
\begin{itemize}[leftmargin=*]
    \item Section~\ref{supp:segalign}. Details on identifying \bbstep instructions and aligning \bbsteps with expert trajectories. (\S~4.3 and \S~4.4 of the main text)
    \item Section~\ref{supp:impl}. Implementation details of the navigation agent, reward function used in RL and optimization hyper-parameters. (\S~4.4 of the main text)
    \item Section~\ref{supp:exp}. Additional experimental results, including in-domain \& transfer results of different dataset trained models, sanity check of our reimplementation, and extra analysis of \bbwalk. (\S~5.1 and \S~5.2 of the main text)
\end{itemize}

\section{Details on \bbstep Identification and Trajectory Alignments}
\label{supp:segalign}
In this section, we describe the details of how \bbsteps are identified in the annotated natural language instructions and how expert trajectory data are segmented to align with \bbstep instructions.

\subsection{Identify \bbsteps}
\label{supp:segalign:seg}
We identify the navigable \bbsteps from the natural language instructions of \rtwor, \rfourr, \rsixr and \reightr, based on the following 6 steps:

\begin{enumerate}[leftmargin=*]
	\item \textbf{Split sentence and chunk phrases.} We split the instructions by periods. For each sentence, we perform POS tagging using the SpaCy~\cite{spacy2} package to locate and chunk all plausible noun phrases and verb phrases.
	\item \textbf{Curate noun phrases.} We curate noun phrases by removing the stop words (\ie, the, for, from \etc) and isolated punctuations among them and lemmatizing each word of them. The purpose is to collect a concentrated set of semantic noun phrases that contain potential visual objects.
	\item \textbf{Identify ``landmark words''.} Next, given the set of candidate visual object words, we filter out a blacklist of words that either do  not correspond to any visual counterpart or are mis-classified by the SpaCy package. The word blacklist includes: 
	\begin{itemize}[leftmargin=*]
        \item[] \texttt{end, 18 inch, head, inside, forward, position, ground, home, face, walk, feet, way, walking, bit, veer, 've, next, stop, towards, right, direction, thing, facing, side, turn, middle, one, out, piece, left, destination, straight, enter, wait, don't, stand, back, round}
     \end{itemize}
     We use the remaining noun phrases as the ``landmark words'' of the sentences. Note that this step identifies the ``landmark words'' for the later procedure which aligns \bbsteps and expert trajectories.
	\item \textbf{Identifying verb phrases.} Similarly, we use a verb blacklist to filter out verbs that require no navigational actions of the agent. The blacklist includes: \texttt{make, turn, face, facing, veer}.
	\item \textbf{Merge non-actionable sentences.} We merge the sentence without landmarks and verbs into the next sentence, as it is likely not actionable. 
	\item \textbf{Merge stop sentences.} There are sentences that only describe the stop condition of a navigation action, which include verb-noun compositions indicating the stop condition. We detect the sentences starting with \texttt{wait, stop, there, remain, you will see} as the sentences that only describe the stop condition and merge them to the previous sentence. Similarly, we detect sentences starting with \texttt{with, facing} and merge them to the next sentence.
\end{enumerate}

After applying the above 6 heuristic rules to the language instruction, we obtain chunks of sentences that describes the navigable \bbsteps of the whole task (\ie, a sequence of navigational sub-goals.).

\subsection{Align Expert Trajectories with identified \bbsteps} 
\label{supp:segalign:ldmk}

In the previous section, we describe the algorithm for identifying \bbstep instructions from the original natural language instructions of the dataset. Now we are going to describe the procedure of aligning \bbsteps with the expert trajectories, which segments the expert trajectories according to the \bbsteps to create the training data for the learning pipeline of our \bbwalk agent. Note that during the training, our \ourmethod \emph{{does not rely on the existence of ground-truth alignments}} between the (micro)instructions and \bbsteps trajectories.

\paragraph{Main Idea} The main idea here is to: 1) perform visual landmark classification to produce confidence scores of landmarks for each visual state $\bss$ along expert trajectories; 2) use the predicted landmark scores and the ``landmark words'' in \bbsteps to guide the alignment between the expert trajectory and \bbsteps. To achieve this, we train a visual landmark classifier with weak supervision --- trajectory-wise existence of landmark objects. Next, based on the predicted landmark confidence scores, we use dynamic programming (DP) to chunk the expert trajectory into segments and assign the segments to the \bbsteps.

\paragraph{Weakly Supervised Learning of the Landmark Classifier}
Given the pairs of aligned instruction and trajectories $\paren{\rmX, \rmY}$ from the original dataset, we train a landmark classifier to detect landmarks mentioned in the instructions. We formulate it as a multi-label classification problem that asks a classifier $f_{\;\textsc{ldmk}}\paren{\bss_t;\calO}$ to predict all the landmarks $\calO_{\rmX}$ of the instruction $\rmX$ given the corresponding trajectory $\rmY$. Here, we denotes all possible landmarks from the entire dataset to be $\calO$, and the landmarks of a specific instruction $\rmX$ to be $\calO_{\rmX}$. Concretely, we first train a convolutional neural network (CNN) based on the visual state features $\bss_t$ to independently predict the existence of landmarks at every time step, then we aggregate the predictions across all time steps to get trajectory-wise logits $\bspsi$ via max-pooling over all states of the trajectory. 
\begin{align}
\small
\bspsi = \texttt{max}\braces{f_{\;\textsc{ldmk}}\paren{\bss_t; \calO } \mid t=1,\ldots,\abs{\rmY}} \nonumber
\end{align}
Here $f_{\;\textsc{ldmk}}$ denotes the independent state-wise landmark classifier, and $\bspsi$ is the logits before normalization for computing the landmark probability. For the specific details of $f_{\;\textsc{ldmk}}$, we input the $6\times6$ panorama visual feature (\ie ResNet-152 feature) into a two-layer CNN (with kernel size of 3, hidden dimension of 128 and ReLU as non-linearity layer) to produce feature activation with spatial extents, followed by a global averaging operator over spatial dimensions and a multi-layer perceptron (2-layer with hidden dimension of 512 and ReLU as non-linearity layer) that outputs the state-wise logits for all visual landmarks $\calO$. We then max pool all the state-wise logits along the trajectory and compute the loss using a trajectory-wise binary cross-entropy between the ground-truth landmark label (of existence) and the prediction.

\paragraph{Aligning \bbsteps and Trajectories with Visual Landmarks} Now, sppose we have a sequence of \bbstep instructions $\rmX = \braces{\bsx_{m},\; m=1,\ldots,\rmM}$, and its expert trajectory $\rmY = \braces{\bss_t,\; t=1,\ldots,{\abs{\rmY}}}$, we can compute the averaged landmark score for the landmarks $\calO_{\bsx_m}$ that exists in this sub-task instruction $\bsx_m$ on a single state $\bss_t$: 
\begin{align}
\small
\Psi\paren{t, m} = \frac{ \mathbf{1}\brackets{\bso_m \in \calO_{\bsx_m}}^\top  f_{\;\textsc{ldmk}}\paren{\bss_t;\calO} }{\abs{\calO_{\bsx_m}}} \nonumber
\end{align}
Here $\mathbf{1}\brackets{\bso_m \in \calO}$ represents the one-hot encoding of the landmarks that exists in the \bbstep $\bsx_m$, and $\abs{\calO_{\bsx_m}}$ is the total number of existed landmarks. We then apply dynamic programming (DP) to solve the trajectory segmentation specified by the following Bellman equation (in a recursive form).
\begin{align}
\small
\Phi\paren{t, m} =
\begin{cases}
	\Psi(t, m), &\mbox{\text{if \;}$t=1$} \\[4pt]
	\Psi(t, m)\; + \\
	\max\limits_{i \in \{1, \ldots, t-1\}} \big\{ \Phi(i, \;m-1) \big\}, &\mbox{\text{otherwise}}
\end{cases} \nonumber
\end{align}
Here, $\Phi\paren{t, m}$ represents the maximum potential of choosing the state $\bss_t$ as the end point of the \bbstep instruction $\bsx_m$. Solving this DP leads to a set of correspondingly segmented trajectories $\rmY = \braces{\bsy_m,\; m = 1,\ldots,\rmM}$, with $\bsy_m$ being the $m$-th \bbstep sub-trajectory. 

\section{Implementation details}
\label{supp:impl}

\subsection{Navigation Agent Configurations} 
\label{supp:impl:agent}
\begin{figure*}[t]
\centering
\includegraphics[width=1\textwidth]{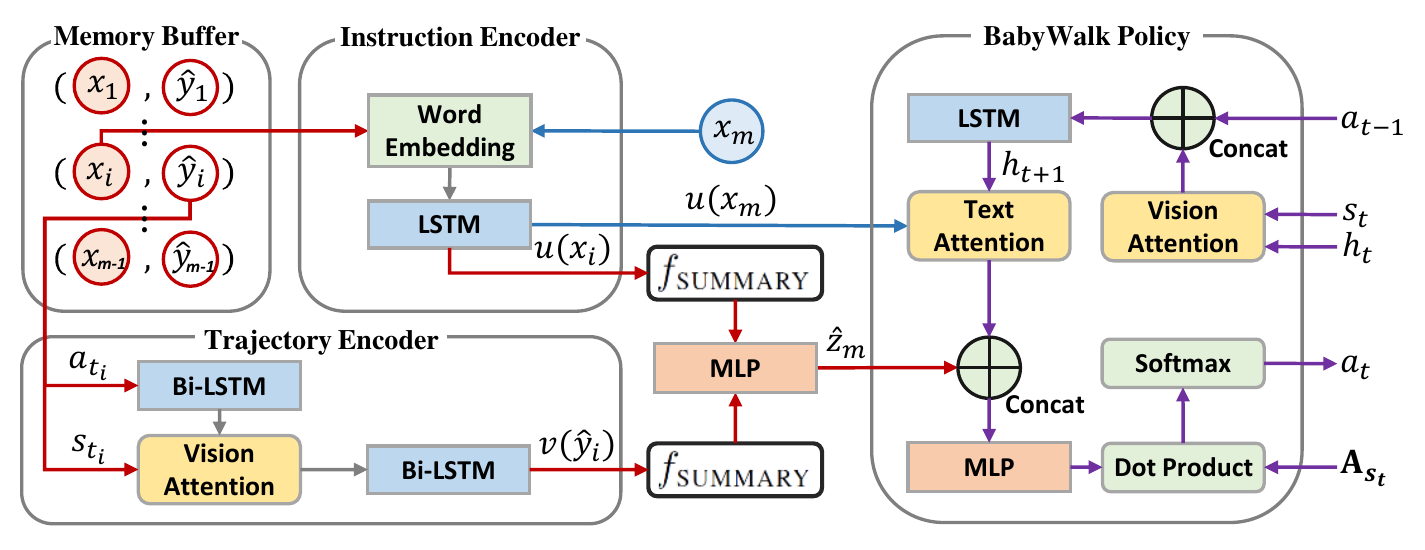}
\caption{
    Our network architecture at the $m$-\textit{th} \bbstep sub-task.
    \textcolor{red}{\bf Red line} represents the procedure of encoding context variable $\bsz_m$ via summarizing the \bbstep trajectory $\summary(v(\hat{\bsy}_1),\ldots,v(\hat{\bsy}_{m-1}))$ and the corresponding (micro)instruction $\summary(u(\bsx_1),\ldots,u(\bsx_{m-1}))$ in the memory buffer.
    \textcolor{blue}{\bf Blue line} represents the procedure of encoding the (micro)instruction $u(\bsx_m)$ of the current \bbstep. \textcolor{darkviolet}{\bf Purple line} represents the detailed decision making process of our \ourmethod policy ($\boldsymbol{\rmA}_{\bss_t}$ is denoted as the set of navigable directions at $\bss_t$ as defined by \citet{fried2018speaker})}
\label{fig:network}
\end{figure*}

Figure~\ref{fig:network} gives an overview of the unrolled version of our full navigation agent.

\paragraph{Panoramic State-Action Space~\cite{fried2018speaker}}
We set up the states $\bss_t$ as the stacked visual feature of agent-centric panoramic views in 12 headings $\times$ 3 elevations with 30 degree intervals. The visual feature of each view is a concatenation of the ResNet-152 feature vector of size 2048 and the orientation feature vector of size 128 (The 4-dimensional orientation feature $[\sin(\phi); \cos(\phi); \sin(\omega); \cos(\omega)]$ are tiled 32 times). We use similar single-view visual feature of size 2176 as our action embeddings.

\paragraph{Encoders}
Instruction encoder $u(\cdot)$ for the instructions is a single directional LSTM with hidden size 512 and a word embedding layer of size 300 (initialized with GloVE embedding~\cite{pennington2014glove}). We use the same encoder for encoding the past experienced and the current executing instruction. Trajectory encoder $v(\cdot)$ contains two separate bidirectional LSTMs (Bi-LSTM), both with hidden size 512. The first Bi-LSTM encodes $\bsa_{t_i}$ and outputs a hidden state for each time step $t_i$. Then we attends the hidden state to the panoramic view $\bss_{t_i}$ to get a state feature of size 2176 for each time step. The second Bi-LSTM encoders the state feature. We use the trajectory encoder just for encoding the past experienced trajectories.

\paragraph{\ourmethod Policy}
The \ourmethod policy network consists of one LSTM with two attention layers and an action predictor. First we attend the hidden state to the panoramic view $\bss_t$ to get state feature of size 2176. The state feature is concatenated with the previous action embedding as a variable to update the hidden state using a LSTM with hidden size 512. The updated hidden state is then attended to the context variables (output of $u(\cdot)$). For the action predictor module, we concatenate the output of text attention layer with the summarized past context $\hat{\bsz}_m$ in order to get an action prediction variable. We then get the action prediction variable through a 2-layer MLP and make a dot product with the navigable action embeddings to retrieve the probability of the next action.

\paragraph{Model Inference} During the inference time, the \ourmethod policy only requires running the heuristic \bbstep identification on the test-time instruction. No need for oracle \bbstep trajectory during this time as the \ourmethod agent is going to roll out for each \bbstep by itself.

\subsection{Details of Reward Shaping for RL} 
\label{supp:impl:rl}

As mentioned in the main text, we learn policy via optimizing the \textit{Fidelity-oriented reward}~\cite{jain2019stay}. Now we give the complete details of this reward function. Suppose the total number of roll out steps is $\rmT = \sum_{i=1}^{\rmM}\abs{\hat{\bsy}_i}$, we would have the following form of reward function:
\begin{equation}
\small
r(\bss_t, \bsa_t) =
    \begin{cases}
        0, & \text{if $t < \rmT$} \\
        \textsc{sr}(\rmY, \hat{\rmY} ) + \textsc{cls}(\rmY, \hat{\rmY}), & \text{if $t = \rmT$}
    \end{cases}
    \nonumber
\end{equation}
Here, $\hat{\rmY} = \hat{\bsy}_1\oplus\ldots\oplus\hat{\bsy}_{\rmM}$ represents the concatenation of \bbstep trajectories produced by the navigation agent (and we note $\oplus$ as the concatenation operation). 

\subsection{Optimization Hyper-parameters} 
\label{supp:impl:hyper}
For each \bbstep task, we set the maximal number of steps to be 10, and truncate the corresponding \bbstep instruction length to be 100. During both the imitation learning and the curriculum reinforcement learning procedures, we fix the learning rate to be 1e-4.
In the imitation learning, the mini-batch size is set to be 100. In the curriculum learning, we reduce the mini-batch size as curriculum increases to save memory consumption. For the 1\textit{st}, 2\textit{nd}, 3\textit{rd} and 4\textit{th} curriculum, the mini-batch size is set to be 50, 32, 20, and 20 respectively.
During the learning, we pre-train our \ourmethod model for 50000 iterations using the imitation learning as a warm-up stage. Next, in each lecture (up to 4) of the reinforcement learning (RL), we train the \ourmethod agent for an additional 10000 iterations, and select the best performing model in terms of \textsc{sdtw} to resume the next lecture. For executing each instruction during the RL, we sample 8 navigation episodes before performing any back-propagation. For each learning stage, we use separate Adam optimizers to optimize for all the parameters.
Meanwhile, we use the L2 weight decay as the regularizer with its coefficient set to be 0.0005. In the reinforcement learning, the discounted factor $\gamma$ is set to be 0.95.  

\section{Additional Experimental Results}
\label{supp:exp}

In this section, we describe a comprehensive set of evaluation metrics and then show transfer results of models trained on each dataset, with all metrics. We provide additional analysis studying the effectiveness of template based \bbstep identification. Finally we present additional qualitative results.

\paragraph{Complete set of Evaluation Metrics.} We adopt the following set of metrics: 
\begin{itemize}[leftmargin=*]
    \item \emph{Path Length (\textsc{pl})} is the length of the agent's navigation path.
    \item \emph{Navigation Error (\textsc{ne})} measures the distance between the goal location and final location of the agent's path. 
    \item \emph{Success Rate (\textsc{sr})} that measures the average rate of the agent stopping within a specified distance near the goal location~\cite{anderson2018vision}
    \item \emph{Success weighted by Path Length (\textsc{spl})}~\cite{anderson2018vision} measures the success rate weighted by the inverse trajectory length, to penalize very long successful trajectory.
    \item \emph{Coverage weighted by Length Score (\textsc{cls})}~\cite{jain2019stay} that measures the fidelity of the agent's path to the reference, weighted by the length score, and the newly proposed
    \item \emph{Normalized Dynamic Time Warping (\textsc{ndtw})} that measures in more fine-grained details, the spatio-temporal similarity of the paths by the agent and the human expert~\cite{magalhaes2019effective}. 
    \item \emph{Success rate weighted normalized Dynamic Time Warping (\textsc{sdtw})} that further measures the spatio-temporal similarity of the paths weighted by the success rate~\cite{magalhaes2019effective}. \textsc{cls}, \textsc{ndtw} and \textsc{sdtw} measure explicitly the agent's ability to follow instructions and in particular, it was shown that \textsc{sdtw} corresponds to human preferences the most. 
\end{itemize}

\subsection{Sanity Check between Prior Methods and Our Re-implementation}
\label{supp:exp:sanity}

\input{supp_tab_sanity}

As mentioned in the main text, we compare our re-implementation and originally reported results of baseline methods on the \rtwor datasets, as Table~\ref{supp:tab:sanity}. We found that the results are mostly very similar, indicating that our re-implementation are reliable.

\subsection{Complete Curriculum Learning Results}
\label{supp:exp:crl}

We present the curriculum learning results with all evaluation metrics in Table~\ref{supp:tab:curriculum}.

\input{supp_tab_crl}

\subsection{Results of \bbstep Identification}
\label{supp:exp:seg}

We present an additional analysis comparing different \bbstep identification methods. We compare our template-based \bbstep identification with a simple method that treat each sentence as an \bbstep (referred as sentence-wise), both using the complete \bbwalk model with the same training routine. The results are shown in the Table~\ref{supp:tab:seg_rules}. Generally speaking, the template based \bbstep identification provides a better performance. 

\input{supp_tab_seg}

\input{supp_tab_indomain}

\subsection{In-domain Results of Models Trained on Instructions with Different lengths}
\label{supp:exp:indomain}

As mentioned in the main text, we display all the in-domain results of navigation agents trained on \rtwor, \rfourr, \rsixr, \reightr, respectively. The complete results of all different metrics are included in the Table~\ref{supp:tab:indomain}. We note that our \bbwalk agent consistently outperforms baseline methods on each dataset. It is worth noting that on \rfourr, \rsixr and \reightr datasets, \textsc{rcm(goal)}$^+$ achieves better results in \textsc{spl}. This is due to the aforementioned fact that they often take short-cuts to directly reach the goal, with a significantly short trajectory. As a consequence, the success rate weighted by inverse path length is high.

\subsection{Transfer Results of Models Trained on Instructions with Different lengths}
\label{supp:exp:transfer}

\input{supp_tab_transfer}

For completeness, we also include all the transfer results of navigation agents trained on \rtwor, \rfourr, \rsixr, \reightr, respectfully. The complete results of all different metrics are included in the Table~\ref{supp:tab:transfer}. According to this  table, we note that models trained on \reightr can achieve the best overall transfer learning performances. This could because of the fact that \reightr trained model only needs to deal with interpolating to shorter ones, rather than extrapolating to longer instructions, which is intuitively an easier direction.

\subsection{Additional Qualitative Results}
\label{supp:exp:qualitative}

We present more qualitative result of various VLN agents as Fig~\ref{fig:sup:viz:path}. It seems that \ourmethod can produce trajectories that align better with the human expert trajectories.

\input{supp_path_viz}

%% file: supp_tab_sanity.tex
\begin{table}[h]
    \centering
	\tabcolsep 2pt
    \resizebox{1.0\linewidth}{!}{ \small
	    \begin{tabular}{@{\;}l@{\;\;}cccc @{\;}}
	        \addlinespace
	        \toprule
	        Data Splits & \multicolumn{4}{c}{\rtwor Validation Unseen} \\
	        Perf. Measures & \textsc{pl} & \textsc{ne}$\downarrow$ & \textsc{sr}$\uparrow$ & \textsc{spl}  \\
	        \midrule
	        \multicolumn{5}{@{\;}l}{\it \footnotesize{Reported Results} } \\
			\textsc{seq2seq}~\cite{fried2018speaker} & - &  7.07 & 31.2 & - \\
			\textsc{sf$^{+}$}~\cite{fried2018speaker}  & - &  6.62 & 35.5 & - \\
			\textsc{rcm$^{+}$}~\cite{wang2019reinforced} & 14.84 & 5.88 & 42.5 & - \\
            \textsc{regretful$^{+\star}$}~\cite{ma2019regretful}  & - & 5.32 & 50.0 & 41.0 \\
            \textsc{fast$^{+\star}$}~\cite{ke2019tactical} & 21.17 & 4.97 & 56.0 & 43.0  \\
	        \midrule
	        \multicolumn{5}{@{\;}l}{\it \footnotesize{Re-implemented Version} } \\
			\textsc{seq2seq}      & 15.76 & 6.71 & 33.6 & 25.5 \\
			\textsc{sf$^{+}$}   & 15.55 & 6.52 & 35.8 & 27.6 \\
			\textsc{rcm$^{+}$}  & 11.15 & 6.18 & 42.4 & 38.6 \\ 
			\textsc{regretful$^{+\star}$} & 13.74 & 5.38 & 48.7 & 39.7 \\
			\textsc{fast$^{+\star}$} & 20.45 & 4.97 & 56.6 & 43.7 \\
	     \bottomrule
	    \end{tabular}
	    }
    \caption{
    	\textbf{Sanity check} of model trained on R2R and evaluated on its validation unseen split ($^{+}$: pre-trained with data augmentation; $\star$:reimplemented or readapted from the original authors’ released code).
    }
    \label{supp:tab:sanity}
\end{table}

%% file: supp_tab_crl.tex
\begin{table}[ht]
    \centering
	\tabcolsep 4pt
    \resizebox{0.9\linewidth}{!}
    {
	    \begin{tabular}{@{\;} cl @{\;\;} cc@{\quad} >{\columncolor[gray]{0.95}}c>{\columncolor[gray]{0.95}}c>{\columncolor[gray]{0.95}}c>{\columncolor[gray]{0.95}}c @{\;}}
	        \toprule
	        \addlinespace
	        & & & & \multicolumn{4}{c}{ \textsc{il}+ \textsc{crl} w/ \textsc{lecture} \# } \\ \cmidrule{5-8}
	        \rotatebox[origin=c]{72}{Datasets} & 
	        \multicolumn{1}{c}{ \rotatebox[origin=c]{72}{Metrics} } & 
			\rotatebox[origin=c]{72}{\textsc{il}} &
	        \rotatebox[origin=c]{72}{\textsc{il}+\textsc{rl}} &
	        \rotatebox[origin=c]{72}{1\textit{st}} &
	        \rotatebox[origin=c]{72}{2\textit{nd}} &
	        \rotatebox[origin=c]{72}{3\textit{rd}} &
	        \rotatebox[origin=c]{72}{4\textit{th}} \\[12pt]
			\midrule \multirow{7}{*}{\rotatebox[origin=c]{72}{$\rtwor$}}
			& \textsc{pl} & 22.4 & 12.0 & 11.6 & 13.2 & 10.6 & 9.6 \\
			& \textsc{ne}$\downarrow$ & 6.8 & 7.1 & 6.8 & 6.8 & 6.7 & \bf 6.6 \\
			& \textsc{sr}$\uparrow$ & 28.1 & 29.8 & 29.9 & 33.2 & 32.2 & \bf 34.1 \\
			& \textsc{spl}$\uparrow$ & 15.7 & 24.3 & 24.9 & 26.6 & 27.5 & \bf 30.2 \\
			& \textsc{cls}$\uparrow$ & 28.9 & 46.2 & 46.6 & 47.2 & 48.1 & \bf 50.4 \\
			&\textsc{ndtw}$\uparrow$ & 30.6 & 43.8 & 42.5 & 41.0 & 47.7 & \bf 50.0 \\
			& \textsc{sdtw}$\uparrow$ & 16.5 & 23.2 & 23.1 & 24.3 & 25.7 & \bf 27.8 \\
			\midrule \multirow{7}{*}{\rotatebox[origin=c]{72}{$\rfourr$}}
			& \textsc{pl} & 43.4 & 22.8 & 23.9 & 25.5 & 21.4 & 19.0 \\
			& \textsc{ne}$\downarrow$ & 8.4 & 8.6 & 8.5 & 8.4 & \bf 8.0 & 8.2 \\
			& \textsc{sr}$\uparrow$ & 24.7 & 25.0 & 24.1 & 26.7 & \bf 27.9 & 27.3 \\
			& \textsc{spl}$\uparrow$ & 8.2 & 11.2 & 11.0 & 12.3 & 13.7 & \bf 14.7 \\
			& \textsc{cls}$\uparrow$ & 27.9 & 45.5 & 44.8 & 45.9 & 47.4 & \bf 49.4 \\
			&\textsc{ndtw}$\uparrow$ & 24.3 & 34.4 & 32.8 & 33.7 & 38.4 & \bf 39.6 \\
			& \textsc{sdtw}$\uparrow$ & 11.1 & 13.6 & 13.5 & 15.2 & 17.0 & \bf 17.3 \\
			\midrule \multirow{7}{*}{\rotatebox[origin=c]{72}{$\rsixr$}}
			& \textsc{pl} & 68.8 & 35.3 & 37.0 & 40.6 & 33.2 & 28.7 \\
			& \textsc{ne}$\downarrow$ & 9.4 & 9.5 & 9.4 & 9.4 & \bf 8.9 & 9.2 \\
			& \textsc{sr}$\uparrow$ & 22.7 & 23.7 & 21.9 & 23.4 & 24.7 & \bf 25.5 \\
			& \textsc{spl}$\uparrow$ & 4.2 & 7.2 & 6.4 & 6.8 & 8.1 & \bf 9.2 \\
			& \textsc{cls}$\uparrow$ & 24.4 & 43.0 & 41.8 & 42.3 & 44.2 & \bf 47.2 \\
			&\textsc{ndtw}$\uparrow$ & 17.8 & 28.1 & 26.0 & 26.9 & 30.9 & \bf 32.7 \\
			& \textsc{sdtw}$\uparrow$ & 7.7 & 10.8 & 9.7 & 11.0 & 12.7 & \bf 13.6 \\\midrule \multirow{7}{*}{\rotatebox[origin=c]{72}{$\reightr$}}
			& \textsc{pl} & 93.1 & 47.5 & 50.0 & 55.3 & 45.2 & 39.9 \\
			& \textsc{ne}$\downarrow$ & 10.0 & 10.2 & 10.2 & 10.1 & \bf 9.3 & 10.1 \\
			& \textsc{sr}$\uparrow$ & 21.9 & 21.4 & 20.4 & 22.1 & \bf 23.1 & 23.1 \\
			& \textsc{spl}$\uparrow$ & 4.3 & 6.1 & 5.5 & 6.1 & 6.8 & \bf 7.4 \\
			& \textsc{cls}$\uparrow$ & 24.1 & 42.1 & 41.0 & 41.5 & 43.9 & \bf 46.0 \\
			&\textsc{ndtw}$\uparrow$ & 15.5 & 24.6 & 22.9 & 23.8 & 27.7 & \bf 28.2 \\
			& \textsc{sdtw}$\uparrow$ & 6.4 & 8.3 & 7.9 & 9.2 & 10.5 & \bf 11.1 \\
			\midrule \multirow{7}{*}{\rotatebox[origin=c]{72}{Average}}
			& \textsc{pl} & 51.8 & 26.8 & 27.9 & 30.6 & 25.1 & 22.1 \\
			& \textsc{ne}$\downarrow$ & 8.5 & 8.7 & 8.5 & 8.5 & \bf 8.1 & 8.3 \\
			& \textsc{sr}$\uparrow$ & 24.7 & 25.5 & 24.6 & 27.0 & 27.5 & \bf 28.1 \\
			& \textsc{spl}$\uparrow$ & 8.6 & 13.1 & 12.9 & 13.9 & 15.1 & \bf 16.5 \\
			& \textsc{cls}$\uparrow$ & 26.6 & 44.5 & 43.9 & 44.6 & 46.2 & \bf 48.6 \\
			&\textsc{ndtw}$\uparrow$ & 23.0 & 33.9 & 32.2 & 32.4 & 37.4 & \bf 39.0 \\
			& \textsc{sdtw}$\uparrow$ & 11.0 & 14.8 & 14.4 & 15.7 & 17.3 & \bf 18.4 \\

	        \addlinespace
	     \bottomrule
	    \end{tabular}
	}
    \caption{
        Ablation on \ourmethod after each learning stage (trained on \rfourr).
    }
    \label{supp:tab:curriculum}
    \vskip -1em
\end{table}

%% file: supp_tab_seg.tex
\begin{table}[ht]
    \centering
	\tabcolsep 4pt
    \resizebox{0.95\linewidth}{!}
    {
	    \begin{tabular}{@{\;} ll @{\quad\quad} cc @{\;}}
	        \toprule
	        \addlinespace
	        \rotatebox[origin=c]{0}{Datasets} & 
	        \rotatebox[origin=c]{0}{Metrics} & 
	        \rotatebox[origin=c]{0}{Sentence-wise} &
			\rotatebox[origin=c]{0}{Template based} \\
			\midrule \multirow{7}{*}{\rotatebox[origin=c]{0}{$\rtwor$}}
			& \textsc{pl} & 10.3 & 9.6 \\
			& \textsc{ne}$\downarrow$ & 6.8 & \bf 6.6 \\
			& \textsc{sr}$\uparrow$ & 28.7 & \bf 34.1 \\
			& \textsc{spl}$\uparrow$ & 24.9 & \bf 30.2 \\
			& \textsc{cls}$\uparrow$ & 48.3 & \bf 50.4 \\
			&\textsc{ndtw}$\uparrow$ & 43.6 & \bf 50.0 \\
			& \textsc{sdtw}$\uparrow$ & 22.4 & \bf 27.8 \\
			\midrule \multirow{7}{*}{\rotatebox[origin=c]{0}{$\rfourr$}}
			& \textsc{pl} & 20.9 & 19.0 \\
			& \textsc{ne}$\downarrow$ & \bf 8.2 & 8.2 \\
			& \textsc{sr}$\uparrow$ & 26.3 & \bf 27.3 \\
			& \textsc{spl}$\uparrow$ & 12.7 & \bf 14.7 \\
			& \textsc{cls}$\uparrow$ & 46.4 & \bf 49.4 \\
			&\textsc{ndtw}$\uparrow$ & 35.5 & \bf 39.6 \\
			& \textsc{sdtw}$\uparrow$ & 15.9 & \bf 17.3 \\
			\midrule \multirow{7}{*}{\rotatebox[origin=c]{0}{$\rsixr$}}
			& \textsc{pl} & 32.1 & 28.7 \\
			& \textsc{ne}$\downarrow$ & \bf 9.0 & 9.2 \\
			& \textsc{sr}$\uparrow$ & 22.5 & \bf 25.5 \\
			& \textsc{spl}$\uparrow$ & 7.5 & \bf 9.2 \\
			& \textsc{cls}$\uparrow$ & 44.2 & \bf 47.2 \\
			&\textsc{ndtw}$\uparrow$ & 29.3 & \bf 32.7 \\
			& \textsc{sdtw}$\uparrow$ & 11.1 & \bf 13.6 \\
            \midrule \multirow{7}{*}{\rotatebox[origin=c]{0}{$\reightr$}}
			& \textsc{pl} & 42.9 & 39.9 \\
			& \textsc{ne}$\downarrow$ & \bf 9.8 & 10.1 \\
			& \textsc{sr}$\uparrow$ & 21.2 & \bf 23.1 \\
			& \textsc{spl}$\uparrow$ & 6.3 & \bf 7.4 \\
			& \textsc{cls}$\uparrow$ & 43.2 & \bf 46.0 \\
			&\textsc{ndtw}$\uparrow$ & 25.5 & \bf 28.2 \\
			& \textsc{sdtw}$\uparrow$ & 9.3 & \bf 11.1 \\
			\midrule \multirow{7}{*}{\rotatebox[origin=c]{0}{Average}}
			& \textsc{pl} & 24.2 & 22.1 \\
			& \textsc{ne}$\downarrow$ & \bf 8.3 & 8.3 \\
			& \textsc{sr}$\uparrow$ & 25.2 & \bf 28.1 \\
			& \textsc{spl}$\uparrow$ & 13.8 & \bf 16.5 \\
			& \textsc{cls}$\uparrow$ & 45.9 & \bf 48.6 \\
			&\textsc{ndtw}$\uparrow$ & 34.6 & \bf 39.0 \\
			& \textsc{sdtw}$\uparrow$ & 15.4 & \bf 18.4 \\
	        \addlinespace
	     \bottomrule
	    \end{tabular}
	}
	\caption{
    	\ourmethod Agent performances between different segmentation rules (trained on \rfourr). Refer to text for more details. 
    }
    \label{supp:tab:seg_rules}
    \vskip -1em
\end{table}

%% file: supp_tab_indomain.tex
\begin{table}[ht]
    \centering
	\tabcolsep 1pt
    \resizebox{0.95\linewidth}{!}
    {
	    \begin{tabular}{@{\;} c@{\;\;}l @{\;\;} cccc @{\quad} >{\columncolor[gray]{0.95}}c>{\columncolor[gray]{0.95}}c @{\;}}
	        \toprule
	        \addlinespace
	        \rotatebox[origin=c]{72}{Datasets} & 
	        \multicolumn{1}{c}{\rotatebox[origin=c]{72}{Metrics}} &
	        \rotatebox[origin=c]{72}{\textsc{seq2seq}} &
	        \rotatebox[origin=c]{72}{\textsc{sf$^{+}$}} &
	        \rotatebox[origin=c]{72}{\textsc{rcm(goal)$^{+}$}} &
	        \rotatebox[origin=c]{72}{\textsc{rcm(fidelity)$^{+}$}} &
	        \rotatebox[origin=c]{72}{{\ourmethod}} &
			\rotatebox[origin=c]{72}{{\ourmethod$^+$}} \\
			\midrule \multirow{7}{*}{\rotatebox[origin=c]{72}{$\rtwor\rightarrow\rtwor$}}
			& \textsc{pl} & 15.8 & 15.6 & 11.1 & 10.2 & 10.7 & 10.2 \\
			& \textsc{ne}$\downarrow$ & 6.7 & 6.5 & 6.2 & 6.2 & 6.2 & \bf 5.9 \\
			& \textsc{sr}$\uparrow$ & 33.6 & 35.8 & 42.4 & 42.1 & 42.6 & \bf 43.8 \\
			& \textsc{spl}$\uparrow$ & 25.5 & 27.6 & 38.6 & 38.6 & 38.3 & \bf 39.6 \\
			& \textsc{cls}$\uparrow$ & 38.5 & 39.8 & 52.7 & 52.6 & 52.9 & \bf 54.4 \\
			&\textsc{ndtw}$\uparrow$ & 39.2 & 41.0 & 51.0 & 50.8 & 53.4 & \bf 55.3 \\
			& \textsc{sdtw}$\uparrow$ & 24.9 & 27.2 & 33.5 & 34.4 & 35.7 & \bf 36.9 \\
			\midrule \multirow{7}{*}{\rotatebox[origin=c]{72}{$\rfourr\rightarrow\rfourr$}}
			& \textsc{pl} & 28.5 & 26.1 & 12.3 & 26.4 & 23.8 & 19.0 \\
			& \textsc{ne}$\downarrow$ & 8.5 & 8.3 & 7.9 & 8.4 & \bf 7.9 & 8.2 \\
			& \textsc{sr}$\uparrow$ & 25.7 & 24.9 & 28.7 & 24.7 & \bf 29.6 & 27.3 \\
			& \textsc{spl}$\uparrow$ & 14.1 & 16.0 & \bf 22.1 & 11.6 & 14.0 & 14.7 \\
			& \textsc{cls}$\uparrow$ & 20.7 & 23.6 & 36.3 & 39.2 & 47.8 & \bf 49.4 \\
			&\textsc{ndtw}$\uparrow$ & 20.6 & 22.7 & 31.3 & 31.3 & 38.1 & \bf 39.6 \\
			& \textsc{sdtw}$\uparrow$ & 9.0 & 9.2 & 13.2 & 13.7 & \bf 18.1 & 17.3 \\
			\midrule \multirow{7}{*}{\rotatebox[origin=c]{72}{$\rsixr\rightarrow\rsixr$}}
			& \textsc{pl} & 34.1 & 43.4 & 11.8 & 28.0 & 28.4 & 27.2 \\
			& \textsc{ne}$\downarrow$ & 9.5 & 9.6 & \bf 9.2 & 9.4 & 9.4 & 9.3 \\
			& \textsc{sr}$\uparrow$ & 18.1 & 17.8 & 18.2 & 20.5 & 21.7 & \bf 22.0 \\
			& \textsc{spl}$\uparrow$ & 9.6 & 7.9 & \bf 14.8 & 7.4 & 7.8 & 8.1 \\
			& \textsc{cls}$\uparrow$ & 23.4 & 20.3 & 31.6 & 39.0 & 47.1 & \bf 47.4 \\
			&\textsc{ndtw}$\uparrow$ & 19.3 & 17.8 & 25.9 & 25.8 & 32.6 & \bf 33.4 \\
			& \textsc{sdtw}$\uparrow$ & 6.5 & 5.9 & 7.6 & 9.5 & 11.5 & \bf 11.8 \\
			\midrule \multirow{7}{*}{\rotatebox[origin=c]{72}{$\reightr\rightarrow\reightr$}}
			& \textsc{pl} & 40.0 & 53.0 & 12.4 & 42.3 & 35.6 & 39.1 \\
			& \textsc{ne}$\downarrow$ & 9.9 & 10.1 & 10.2 & 10.7 & \bf 9.6 & 9.9 \\
			& \textsc{sr}$\uparrow$ & 20.2 & 18.6 & 19.7 & 18.2 & \bf 22.3 & 22.0 \\
			& \textsc{spl}$\uparrow$ & 12.4 & 9.8 & \bf 15.4 & 5.3 & 7.3 & 7.0 \\
			& \textsc{cls}$\uparrow$ & 19.8 & 16.3 & 25.7 & 37.2 & 46.4 & \bf 46.4 \\
			&\textsc{ndtw}$\uparrow$ & 15.8 & 13.5 & 19.4 & 21.6 & \bf 29.6 & 28.3 \\
			& \textsc{sdtw}$\uparrow$ & 5.1 & 4.4 & 5.8 & 7.6 & \bf 10.4 & 10.1 \\
	        \addlinespace
	     \bottomrule
	    \end{tabular}
	}
	\caption{
    	\textbf{Indomain results.} Each model is trained on the training set of \rtwor, \rfourr, \rsixr and \reightr datasets, and evaluated on the corresponding unseen validation set ($^+$: pre-trained with data augmentation).
    }
    \label{supp:tab:indomain}
    \vskip -1em
\end{table}

%% file: supp_tab_transfer.tex
\begin{table*}[htp]
    \centering
    \tabcolsep 5pt
    \begin{tabular}{cc}
        \begin{minipage}{0.5\textwidth}
            \input{supp_tab_r2r}
        \end{minipage}
        & 
        \begin{minipage}{0.5\textwidth}
            \input{supp_tab_r4r}
        \end{minipage} \\
        \bf (a) \rtwor trained model & \bf (b) \rfourr trained model \\
        \addlinespace
        \begin{minipage}{0.5\textwidth}
            \input{supp_tab_r6r}
        \end{minipage}
        & 
        \begin{minipage}{0.5\textwidth}
            \input{supp_tab_r8r}
        \end{minipage} \\
        \bf (c) \rsixr trained model & \bf (d) \reightr trained model
    \end{tabular}
    \caption{ \textbf{Transfer results} of \rtwor, \rfourr, \rsixr, \reightr trained model evaluated on their complementary unseen validation datasets ($^+$: pre-trained with data augmentation; $^\star$: reimplemented or readapted from the original authors' released code).}
    \label{supp:tab:transfer}
\end{table*}

%% file: supp_tab_r2r.tex
    \centering
	\tabcolsep 1pt
    \resizebox{0.85\linewidth}{!}
    {
	    \begin{tabular}{@{\;} c@{\;\;}l @{\;} cccccc @{\;\;} >{\columncolor[gray]{0.95}}c>{\columncolor[gray]{0.95}}c @{\;}}
	        \toprule
	        \addlinespace
	        \rotatebox[origin=c]{72}{Datasets} & 
	        \multicolumn{1}{c}{\rotatebox[origin=c]{72}{Metrics}} & \rotatebox[origin=c]{72}{\textsc{seq2seq}} &
	        \rotatebox[origin=c]{72}{\textsc{sf$^{+}$}} &
	        \rotatebox[origin=c]{72}{\textsc{rcm(goal)$^{+}$}} &
	        \rotatebox[origin=c]{72}{\textsc{rcm(fidelity)$^{+}$}} &
	        \rotatebox[origin=c]{72}{\textsc{regretful$^{+\star}$}} &
	        \rotatebox[origin=c]{72}{\textsc{fast$^{+\star}$}} & 
			\rotatebox[origin=c]{72}{{\ourmethod}} &
			\rotatebox[origin=c]{72}{{\ourmethod$^+$}} \\
			\midrule \multirow{7}{*}{\rotatebox[origin=c]{72}{$\rtwor\rightarrow\rfourr$}}
			& \textsc{pl} & 28.6 & 28.9 & 13.2 & 14.1 & 15.5 & 29.7 & 19.5 & 17.9 \\
			& \textsc{ne}$\downarrow$ & 9.1 & 9.0 & 9.2 & 9.3 & \bf 8.4 & 9.1 & 8.9 & 8.9 \\
			& \textsc{sr}$\uparrow$ & 18.3 & 16.7 & 14.7 & 15.2 & 19.2 & 13.3 & \bf 22.5 & 21.4 \\
			& \textsc{spl}$\uparrow$ & 7.9 & 7.4 & 8.9 & 8.9 & 10.1 & 7.7 & \bf 12.6 & 11.9 \\
			& \textsc{cls}$\uparrow$ & 29.8 & 30.0 & 42.5 & 41.2 & 46.4 & 41.8 & 50.3 & \bf 51.0 \\
			&\textsc{ndtw}$\uparrow$ & 25.1 & 25.3 & 33.3 & 32.4 & 31.6 & 33.5 & 38.9 & \bf 40.3 \\
			& \textsc{sdtw}$\uparrow$ & 7.1 & 6.7 & 7.3 & 7.2 & 9.8 & 7.2 & \bf 14.5 & 13.8 \\
			\midrule \multirow{7}{*}{\rotatebox[origin=c]{72}{$\rtwor\rightarrow\rsixr$}}
			& \textsc{pl} & 39.4 & 41.4 & 14.2 & 15.7 & 15.9 & 32.0 & 29.1 & 25.9 \\
			& \textsc{ne}$\downarrow$ & 9.6 & 9.8 & 9.7 & 9.8 & \bf 8.8 & 9.0 & 10.1 & 9.8 \\
			& \textsc{sr}$\uparrow$ & 20.7 & 17.9 & 22.4 & 22.7 & 24.2 & \bf 26.0 & 21.4 & 21.7 \\
			& \textsc{spl}$\uparrow$ & 11.0 & 9.1 & 17.7 & \bf 18.3 & 16.6 & 16.5 & 7.9 & 8.8 \\
			& \textsc{cls}$\uparrow$ & 25.9 & 26.2 & 37.1 & 36.4 & 40.9 & 37.7 & 48.4 & \bf 49.0 \\
			&\textsc{ndtw}$\uparrow$ & 20.5 & 20.8 & 26.6 & 26.1 & 16.2 & 21.9 & 30.8 & \bf 32.6 \\
			& \textsc{sdtw}$\uparrow$ & 7.7 & 7.2 & 8.2 & 8.4 & 6.8 & 8.5 & 11.2 & \bf 11.2 \\
			\midrule \multirow{7}{*}{\rotatebox[origin=c]{72}{$\rtwor\rightarrow\reightr$}}
			& \textsc{pl} & 52.3 & 52.2 & 15.3 & 16.9 & 16.6 & 34.9 & 38.3 & 34.0 \\
			& \textsc{ne}$\downarrow$ & 10.5 & 10.5 & 11.0 & 11.1 & \bf 10.0 & 10.6 & 11.1 & 10.5 \\
			& \textsc{sr}$\uparrow$ & 16.9 & 13.8 & 12.4 & 12.6 & 16.3 & 11.1 & 19.6 & \bf 20.7 \\
			& \textsc{spl}$\uparrow$ & 6.1 & 5.6 & 7.4 & 7.5 & 7.7 & 6.2 & 6.9 & \bf 7.8 \\
			& \textsc{cls}$\uparrow$ & 22.5 & 24.1 & 32.4 & 30.9 & 35.3 & 33.7 & 48.1 & \bf 48.7 \\
			&\textsc{ndtw}$\uparrow$ & 17.1 & 18.2 & 23.9 & 23.3 & 8.1 & 14.5 & 26.7 & \bf 29.1 \\
			& \textsc{sdtw}$\uparrow$ & 4.1 & 3.8 & 4.3 & 4.3 & 2.4 & 2.4 & 9.4 & \bf 9.8 \\
			\midrule \multirow{7}{*}{\rotatebox[origin=c]{72}{Average}}
			& \textsc{pl} & 40.1 & 40.8 & 14.2 & 15.6 & 16.0 & 32.2 & 29.0 & 25.9 \\
			& \textsc{ne}$\downarrow$ & 9.7 & 9.8 & 10.0 & 10.1 & \bf 9.1 & 9.6 & 10.0 & 9.7 \\
			& \textsc{sr}$\uparrow$ & 18.6 & 16.1 & 16.5 & 16.8 & 19.9 & 16.8 & 21.2 & \bf 21.3 \\
			& \textsc{spl}$\uparrow$ & 8.3 & 7.4 & 11.3 & \bf 11.6 & 11.5 & 10.1 & 9.1 & 9.5 \\
			& \textsc{cls}$\uparrow$ & 26.1 & 26.8 & 37.3 & 36.2 & 40.9 & 37.7 & 48.9 & \bf 49.6 \\
			&\textsc{ndtw}$\uparrow$ & 20.9 & 21.4 & 27.9 & 27.3 & 18.6 & 23.3 & 32.1 & \bf 34.0 \\
			& \textsc{sdtw}$\uparrow$ & 6.3 & 5.9 & 6.6 & 6.6 & 6.3 & 6.0 & \bf 11.7 & 11.6 \\
	        \addlinespace
	     \bottomrule
	    \end{tabular}
	}

%% file: supp_tab_r4r.tex
    \centering
	\tabcolsep 1pt
    \resizebox{0.85\linewidth}{!}
    {
	    \begin{tabular}{@{\;} c@{\;\;}l @{\;} cccccc @{\;\;} >{\columncolor[gray]{0.95}}c>{\columncolor[gray]{0.95}}c @{\;}}
	        \toprule
	        \addlinespace
	        \rotatebox[origin=c]{72}{Datasets} & 
	        \multicolumn{1}{c}{\rotatebox[origin=c]{72}{Metrics}} & \rotatebox[origin=c]{72}{\textsc{seq2seq}} &
	        \rotatebox[origin=c]{72}{\textsc{sf$^{+}$}} &
	        \rotatebox[origin=c]{72}{\textsc{rcm(goal)$^{+}$}} &
	        \rotatebox[origin=c]{72}{\textsc{rcm(fidelity)$^{+}$}} &
	        \rotatebox[origin=c]{72}{\textsc{regretful$^{+\star}$}} &
	        \rotatebox[origin=c]{72}{\textsc{fast$^{+\star}$}} & 
			\rotatebox[origin=c]{72}{{\ourmethod}} &
			\rotatebox[origin=c]{72}{{\ourmethod$^+$}} \\
			\midrule \multirow{7}{*}{\rotatebox[origin=c]{72}{$\rfourr\rightarrow\rtwor$}}
			& \textsc{pl} & 16.2 & 17.4 & 10.2 & 17.7 & 20.0 & 26.5 & 12.1 & 9.6 \\
			& \textsc{ne}$\downarrow$ & 7.8 & 7.3 & 7.1 & 6.7 & 7.5 & 7.2 & \bf 6.6 & 6.6 \\
			& \textsc{sr}$\uparrow$ & 16.3 & 22.5 & 25.9 & 29.1 & 22.8 & 25.1 & \bf 35.2 & 34.1 \\
			& \textsc{spl}$\uparrow$ & 9.9 & 14.1 & 22.5 & 18.2 & 14.0 & 16.3 & 28.3 & \bf 30.2 \\
			& \textsc{cls}$\uparrow$ & 27.1 & 29.5 & 44.2 & 34.3 & 32.6 & 33.9 & 48.5 & \bf 50.4 \\
			&\textsc{ndtw}$\uparrow$ & 29.3 & 31.8 & 41.1 & 33.5 & 28.5 & 27.9 & 46.5 & \bf 50.0 \\
			& \textsc{sdtw}$\uparrow$ & 10.6 & 14.8 & 20.2 & 18.3 & 13.4 & 14.2 & 27.2 & \bf 27.8 \\
			\midrule \multirow{7}{*}{\rotatebox[origin=c]{72}{$\rfourr\rightarrow\rsixr$}}
			& \textsc{pl} & 40.8 & 38.5 & 12.8 & 33.0 & 19.9 & 26.6 & 37.0 & 28.7 \\
			& \textsc{ne}$\downarrow$ & 9.9 & 9.5 & 9.2 & 9.3 & 9.5 & 8.9 & \bf 8.8 & 9.2 \\
			& \textsc{sr}$\uparrow$ & 14.4 & 15.5 & 19.3 & 20.5 & 18.0 & 22.1 & \bf 26.4 & 25.5 \\
			& \textsc{spl}$\uparrow$ & 6.8 & 8.4 & \bf 15.2 & 8.5 & 10.6 & 13.7 & 8.1 & 9.2 \\
			& \textsc{cls}$\uparrow$ & 17.7 & 20.4 & 31.8 & 38.3 & 31.7 & 31.5 & 44.9 & \bf 47.2 \\
			&\textsc{ndtw}$\uparrow$ & 16.4 & 18.3 & 23.5 & 23.7 & 23.5 & 23.0 & 30.1 & \bf 32.7 \\
			& \textsc{sdtw}$\uparrow$ & 4.6 & 5.2 & 7.3 & 7.9 & 7.5 & 7.7 & 13.1 & \bf 13.6 \\
			\midrule \multirow{7}{*}{\rotatebox[origin=c]{72}{$\rfourr\rightarrow\reightr$}}
			& \textsc{pl} & 56.4 & 50.8 & 13.9 & 38.7 & 20.7 & 28.2 & 50.0 & 39.9 \\
			& \textsc{ne}$\downarrow$ & 10.1 & 9.5 & 9.5 & 9.9 & 9.5 & \bf 9.1 & 9.3 & 10.1 \\
			& \textsc{sr}$\uparrow$ & 20.7 & 21.6 & 22.8 & 20.9 & 18.7 & \bf 27.7 & 26.3 & 23.1 \\
			& \textsc{spl}$\uparrow$ & 10.4 & 11.8 & \bf 16.9 & 9.0 & 9.2 & 13.7 & 7.2 & 7.4 \\
			& \textsc{cls}$\uparrow$ & 15.0 & 17.2 & 27.6 & 34.6 & 29.3 & 29.6 & 44.7 & \bf 46.0 \\
			&\textsc{ndtw}$\uparrow$ & 13.4 & 15.1 & 19.5 & 21.7 & 19.0 & 17.7 & 27.1 & \bf 28.2 \\
			& \textsc{sdtw}$\uparrow$ & 4.7 & 5.0 & 5.1 & 6.1 & 5.6 & 6.9 & \bf 11.5 & 11.1 \\
			\midrule \multirow{7}{*}{\rotatebox[origin=c]{72}{Average}}
			& \textsc{pl} & 37.8 & 35.6 & 12.3 & 29.8 & 20.2 & 27.1 & 33.0 & 26.1 \\
			& \textsc{ne}$\downarrow$ & 9.3 & 8.8 & 8.6 & 8.6 & 8.8 & 8.4 & \bf 8.2 & 8.6 \\
			& \textsc{sr}$\uparrow$ & 17.1 & 19.9 & 22.7 & 23.5 & 19.8 & 25.0 & \bf 29.3 & 27.6 \\
			& \textsc{spl}$\uparrow$ & 9.0 & 11.4 & \bf 18.2 & 11.9 & 11.3 & 14.6 & 14.5 & 15.6 \\
			& \textsc{cls}$\uparrow$ & 19.9 & 22.4 & 34.5 & 35.7 & 31.2 & 31.7 & 46.0 & \bf 47.9 \\
			&\textsc{ndtw}$\uparrow$ & 19.7 & 21.7 & 28.0 & 26.3 & 23.7 & 22.9 & 34.6 & \bf 37.0 \\
			& \textsc{sdtw}$\uparrow$ & 6.6 & 8.3 & 10.9 & 10.8 & 8.8 & 9.6 & 17.3 & \bf 17.5 \\
        \addlinespace
	     \bottomrule
	    \end{tabular}

	}

%% file: supp_tab_r6r.tex
    \centering
	\tabcolsep 1pt
    \resizebox{0.7\linewidth}{!}
    {
	    \begin{tabular}{@{\;} c@{\;\;}l @{\;\;} cccc @{\quad} >{\columncolor[gray]{0.95}}c>{\columncolor[gray]{0.95}}c @{\;}}
	        \toprule
	        \addlinespace
	        \rotatebox[origin=c]{72}{Datasets} & 
	        \multicolumn{1}{c}{\rotatebox[origin=c]{72}{Metrics}} &
	        \rotatebox[origin=c]{72}{\textsc{seq2seq}} &
	        \rotatebox[origin=c]{72}{\textsc{sf$^{+}$}} &
	        \rotatebox[origin=c]{72}{\textsc{rcm(goal)$^{+}$}} &
	        \rotatebox[origin=c]{72}{\textsc{rcm(fidelity)$^{+}$}} &
	        \rotatebox[origin=c]{72}{{\ourmethod}} &
			\rotatebox[origin=c]{72}{{\ourmethod$^+$}} \\
			\midrule \multirow{7}{*}{\rotatebox[origin=c]{72}{$\rsixr\rightarrow\rtwor$}}
			& \textsc{pl} & 14.5 & 19.4 & 8.1 & 15.5 & 9.4 & 9.2 \\
			& \textsc{ne}$\downarrow$ & 7.7 & 7.1 & 7.6 & 7.5 & 6.8 & \bf 6.8 \\
			& \textsc{sr}$\uparrow$ & 19.3 & 21.9 & 19.6 & 22.6 & \bf 31.3 & 30.6 \\
			& \textsc{spl}$\uparrow$ & 13.3 & 11.6 & 17.2 & 14.1 & \bf 28.3 & 27.8 \\
			& \textsc{cls}$\uparrow$ & 32.1 & 26.2 & 43.2 & 34.3 & 49.9 & \bf 50.0 \\
			&\textsc{ndtw}$\uparrow$ & 31.9 & 30.8 & 39.7 & 32.4 & \bf 49.5 & 49.4 \\
			& \textsc{sdtw}$\uparrow$ & 13.1 & 13.3 & 15.3 & 14.3 & \bf 25.9 & 25.4 \\
			\midrule \multirow{7}{*}{\rotatebox[origin=c]{72}{$\rsixr\rightarrow\rfourr$}}
			& \textsc{pl} & 25.2 & 33.0 & 11.6 & 25.7 & 18.1 & 17.7 \\
			& \textsc{ne}$\downarrow$ & 8.7 & 8.6 & 8.5 & 8.4 & 8.4 & \bf 8.2 \\
			& \textsc{sr}$\uparrow$ & 24.2 & 22.4 & 23.6 & \bf 25.4 & 24.3 & 24.3 \\
			& \textsc{spl}$\uparrow$ & 13.7 & 9.3 & \bf 17.5 & 10.6 & 12.8 & 12.9 \\
			& \textsc{cls}$\uparrow$ & 25.8 & 21.4 & 35.8 & 34.8 & 48.6 & \bf 48.6 \\
			&\textsc{ndtw}$\uparrow$ & 22.9 & 20.6 & 29.8 & 26.5 & 39.0 & \bf 39.4 \\
			& \textsc{sdtw}$\uparrow$ & 9.3 & 7.5 & 10.8 & 11.1 & 15.1 & \bf 15.1 \\
			\midrule \multirow{7}{*}{\rotatebox[origin=c]{72}{$\rsixr\rightarrow\reightr$}}
			& \textsc{pl} & 43.0 & 52.8 & 14.2 & 29.9 & 38.3 & 36.8 \\
			& \textsc{ne}$\downarrow$ & 9.9 & 9.9 & \bf 9.6 & 9.7 & 10.2 & 10.0 \\
			& \textsc{sr}$\uparrow$ & 20.1 & 20.3 & 20.3 & \bf 22.4 & 20.8 & 21.0 \\
			& \textsc{spl}$\uparrow$ & 11.2 & 9.4 & \bf 14.9 & 8.1 & 6.6 & 6.8 \\
			& \textsc{cls}$\uparrow$ & 20.6 & 18.3 & 27.7 & 38.9 & 45.9 & \bf 46.3 \\
			&\textsc{ndtw}$\uparrow$ & 16.3 & 15.2 & 21.9 & 22.2 & 28.4 & \bf 29.3 \\
			& \textsc{sdtw}$\uparrow$ & 5.6 & 5.0 & 6.4 & 6.8 & 9.6 & \bf 9.9 \\
			\midrule \multirow{7}{*}{\rotatebox[origin=c]{72}{Average}}
			& \textsc{pl} & 27.6 & 35.1 & 11.3 & 23.7 & 21.9 & 21.2 \\
			& \textsc{ne}$\downarrow$ & 8.8 & 8.5 & 8.6 & 8.5 & 8.5 & \bf 8.3 \\
			& \textsc{sr}$\uparrow$ & 21.2 & 21.5 & 21.2 & 23.5 & \bf 25.5 & 25.3 \\
			& \textsc{spl}$\uparrow$ & 12.7 & 10.1 & \bf 16.5 & 10.9 & 15.9 & 15.8 \\
			& \textsc{cls}$\uparrow$ & 26.2 & 22.0 & 35.6 & 36.0 & 48.1 & \bf 48.3 \\
			&\textsc{ndtw}$\uparrow$ & 23.7 & 22.2 & 30.5 & 27.0 & 39.0 & \bf 39.4 \\
			& \textsc{sdtw}$\uparrow$ & 9.3 & 8.6 & 10.8 & 10.7 & \bf 16.9 & 16.8 \\

	        \addlinespace
	     \bottomrule
	    \end{tabular}
	}

%% file: supp_tab_r8r.tex
    \centering
	\tabcolsep 1pt
    \resizebox{0.7\linewidth}{!}
    {
	    \begin{tabular}{@{\;} c@{\;\;}l @{\;} cccc @{\quad} >{\columncolor[gray]{0.95}}c>{\columncolor[gray]{0.95}}c @{\;}}
	        \toprule
	        \addlinespace
	        \rotatebox[origin=c]{72}{Datasets} & 
	        \multicolumn{1}{c}{\rotatebox[origin=c]{72}{Metrics}} &
	        \rotatebox[origin=c]{72}{\textsc{seq2seq}} &
	        \rotatebox[origin=c]{72}{\textsc{sf$^{+}$}} &
	        \rotatebox[origin=c]{72}{\textsc{rcm(goal)$^{+}$}} &
	        \rotatebox[origin=c]{72}{\textsc{rcm(fidelity)$^{+}$}} &
	        \rotatebox[origin=c]{72}{{\ourmethod}} &
			\rotatebox[origin=c]{72}{{\ourmethod$^+$}} \\
            \midrule \multirow{7}{*}{\rotatebox[origin=c]{72}{$\reightr\rightarrow\rtwor$}}
			& \textsc{pl} & 13.7 & 19.3 & 7.8 & 17.8 & 9.1 & 9.8 \\
			& \textsc{ne}$\downarrow$ & 7.6 & 7.3 & 8.0 & 8.2 & 6.8 & \bf 6.7 \\
			& \textsc{sr}$\uparrow$ & 18.7 & 23.4 & 14.8 & 19.2 & 30.0 & \bf 32.1 \\
			& \textsc{spl}$\uparrow$ & 13.3 & 12.9 & 12.9 & 10.6 & 27.0 & \bf 28.2 \\
			& \textsc{cls}$\uparrow$ & 32.7 & 26.6 & 37.9 & 28.9 & \bf 49.5 & 49.3 \\
			&\textsc{ndtw}$\uparrow$ & 32.4 & 29.9 & 34.9 & 25.9 & 48.9 & \bf 48.9 \\
			& \textsc{sdtw}$\uparrow$ & 12.7 & 14.5 & 11.1 & 10.5 & 24.6 & \bf 26.2 \\
			\midrule \multirow{7}{*}{\rotatebox[origin=c]{72}{$\reightr\rightarrow\rfourr$}}
			& \textsc{pl} & 23.1 & 31.7 & 11.1 & 32.5 & 17.4 & 19.0 \\
			& \textsc{ne}$\downarrow$ & 8.7 & 8.8 & 8.7 & 9.2 & \bf 8.2 & 8.5 \\
			& \textsc{sr}$\uparrow$ & 23.6 & 21.8 & 23.2 & 21.7 & 24.4 & \bf 24.4 \\
			& \textsc{spl}$\uparrow$ & 15.1 & 10.5 & \bf 18.2 & 7.4 & 12.6 & 12.5 \\
			& \textsc{cls}$\uparrow$ & 24.9 & 20.8 & 32.3 & 29.4 & 48.1 & \bf 48.5 \\
			&\textsc{ndtw}$\uparrow$ & 22.3 & 19.7 & 26.4 & 20.6 & \bf 39.1 & 38.5 \\
			& \textsc{sdtw}$\uparrow$ & 8.8 & 7.7 & 9.3 & 8.4 & 14.9 & \bf 15.2 \\
			\midrule \multirow{7}{*}{\rotatebox[origin=c]{72}{$\reightr\rightarrow\rsixr$}}
			& \textsc{pl} & 30.9 & 42.2 & 11.9 & 39.9 & 26.6 & 29.2 \\
			& \textsc{ne}$\downarrow$ & 9.7 & 9.9 & 9.9 & 10.1 & \bf 9.0 & 9.3 \\
			& \textsc{sr}$\uparrow$ & 15.4 & 14.7 & 14.8 & 20.0 & \bf 22.9 & 22.9 \\
			& \textsc{spl}$\uparrow$ & 8.6 & 6.7 & \bf 11.6 & 5.3 & 8.4 & 7.9 \\
			& \textsc{cls}$\uparrow$ & 22.2 & 18.5 & 29.1 & 33.5 & \bf 46.9 & 46.6 \\
			&\textsc{ndtw}$\uparrow$ & 18.5 & 15.9 & 22.5 & 20.1 & \bf 33.3 & 31.8 \\
			& \textsc{sdtw}$\uparrow$ & 5.5 & 4.7 & 6.0 & 7.8 & \bf 12.1 & 11.8 \\
			\midrule \multirow{7}{*}{\rotatebox[origin=c]{72}{Average}}
			& \textsc{pl} & 22.6 & 31.1 & 10.3 & 30.1 & 17.7 & 19.3 \\
			& \textsc{ne}$\downarrow$ & 8.7 & 8.7 & 8.9 & 9.2 & \bf 8.0 & 8.2 \\
			& \textsc{sr}$\uparrow$ & 19.2 & 20.0 & 17.6 & 20.3 & 25.8 & \bf 26.5 \\
			& \textsc{spl}$\uparrow$ & 12.3 & 10.0 & 14.2 & 7.8 & 16.0 & \bf 16.2 \\
			& \textsc{cls}$\uparrow$ & 26.6 & 22.0 & 33.1 & 30.6 & \bf 48.2 & 48.1 \\
			&\textsc{ndtw}$\uparrow$ & 24.4 & 21.8 & 27.9 & 22.2 & \bf 40.4 & 39.7 \\
			& \textsc{sdtw}$\uparrow$ & 9.0 & 9.0 & 8.8 & 8.9 & 17.2 & \bf 17.7 \\

	        \addlinespace
	     \bottomrule
	    \end{tabular}
	}

%% file: supp_path_viz.tex
\def\VizImageWidth{1.2in}
\def\VizImageHeightLong{1.2in}

\begin{figure*}[htp]
\small
\centering
\tabcolsep 1pt
\begin{tabular}{ccccc}
\toprule
\textsc{human} & \ourmethod & \textsc{rcm} & \textsc{sf} & \textsc{seq2seq} \\
\midrule

\includegraphics[width=\VizImageWidth,height=\VizImageHeightLong]{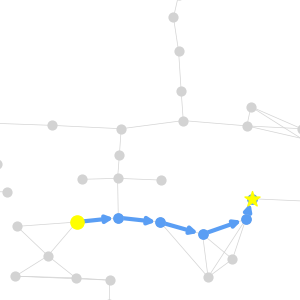} &
\includegraphics[width=\VizImageWidth,height=\VizImageHeightLong]{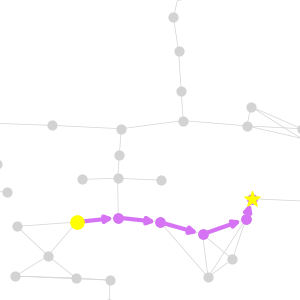} &
\includegraphics[width=\VizImageWidth,height=\VizImageHeightLong]{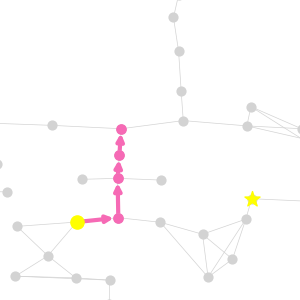} &
\includegraphics[width=\VizImageWidth,height=\VizImageHeightLong]{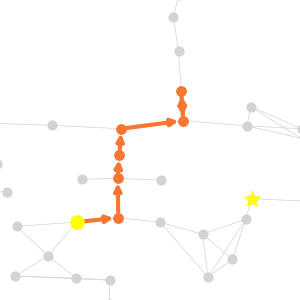} &
\includegraphics[width=\VizImageWidth,height=\VizImageHeightLong]{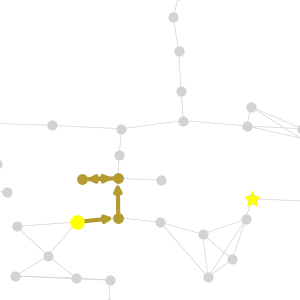} \\

\includegraphics[width=\VizImageWidth,height=\VizImageHeightLong]{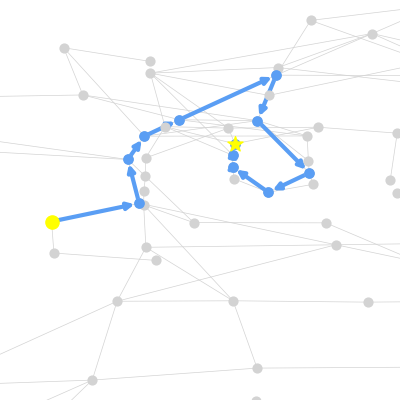} &
\includegraphics[width=\VizImageWidth,height=\VizImageHeightLong]{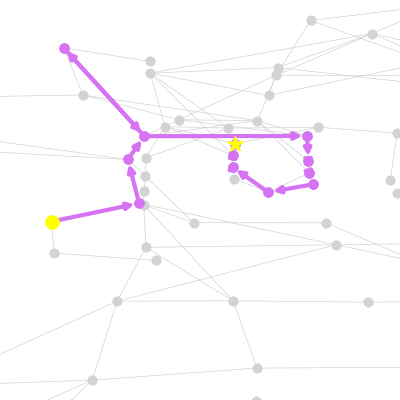} &
\includegraphics[width=\VizImageWidth,height=\VizImageHeightLong]{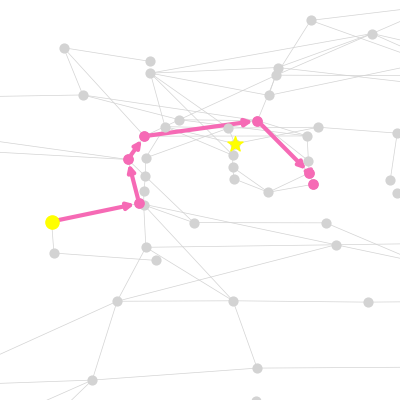} &
\includegraphics[width=\VizImageWidth,height=\VizImageHeightLong]{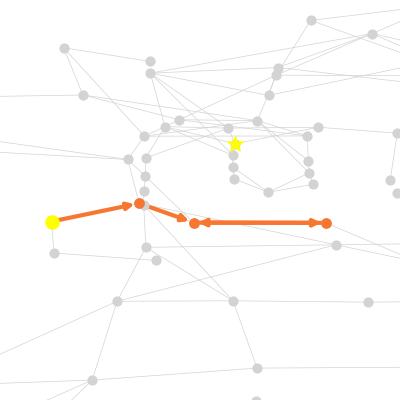} &
\includegraphics[width=\VizImageWidth,height=\VizImageHeightLong]{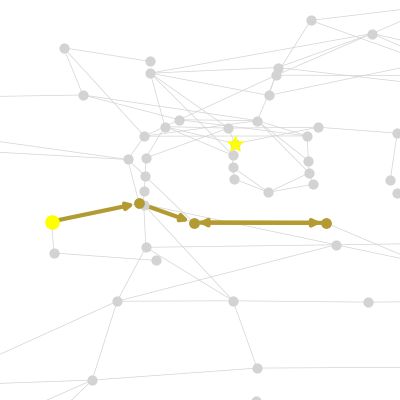} \\

\includegraphics[width=\VizImageWidth,height=\VizImageHeightLong]{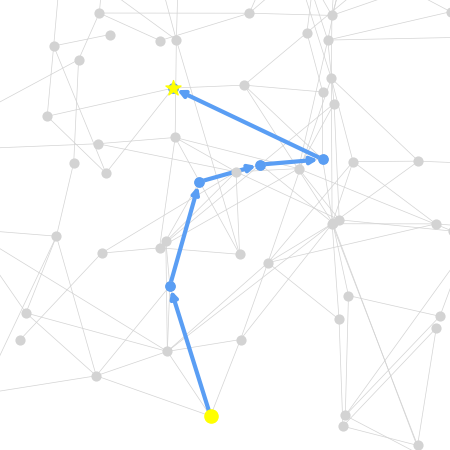} &
\includegraphics[width=\VizImageWidth,height=\VizImageHeightLong]{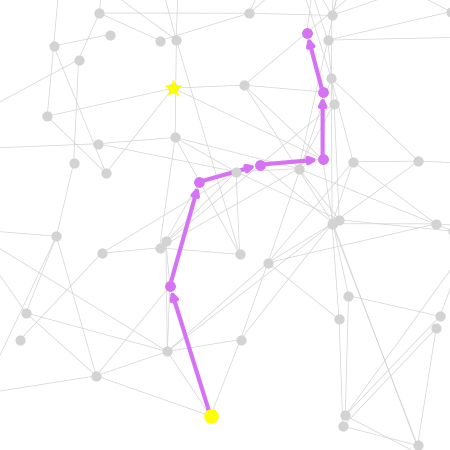} &
\includegraphics[width=\VizImageWidth,height=\VizImageHeightLong]{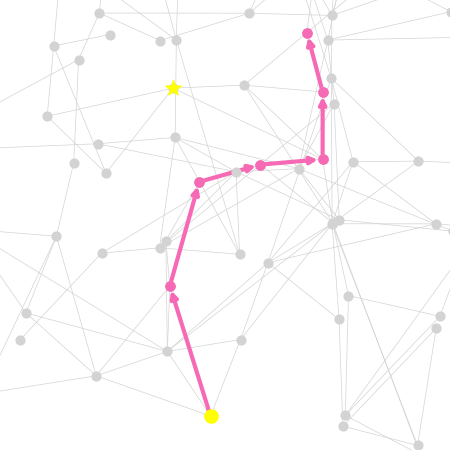} &
\includegraphics[width=\VizImageWidth,height=\VizImageHeightLong]{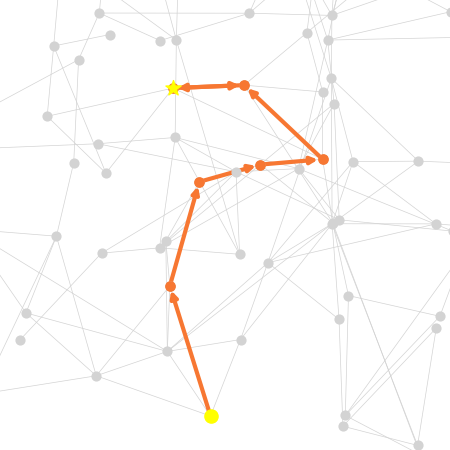} &
\includegraphics[width=\VizImageWidth,height=\VizImageHeightLong]{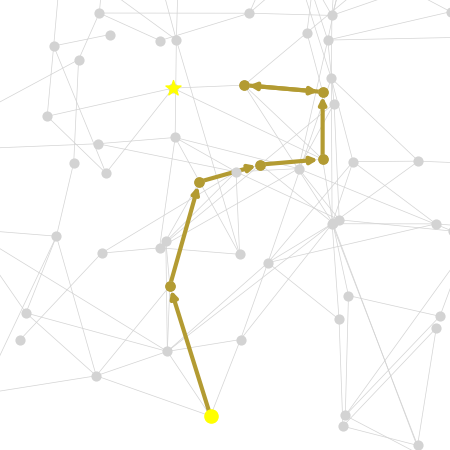} \\

\includegraphics[width=\VizImageWidth,height=\VizImageHeightLong]{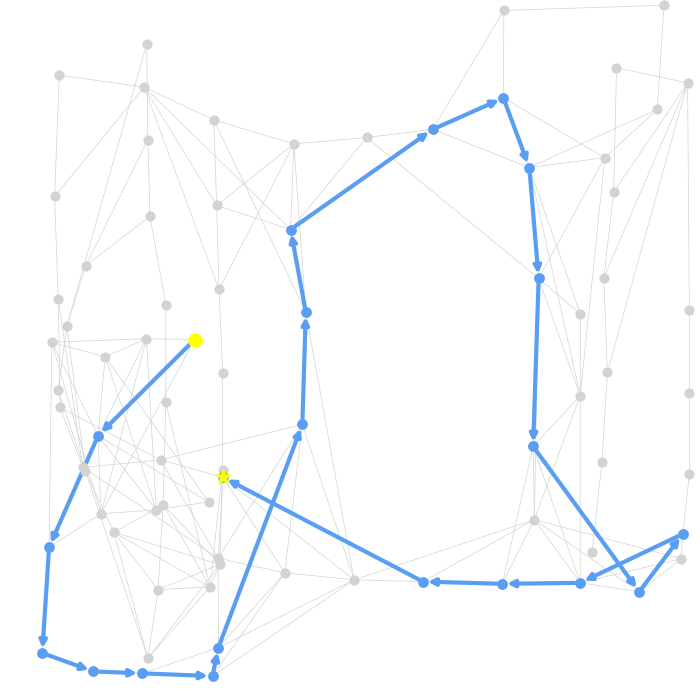} &
\includegraphics[width=\VizImageWidth,height=\VizImageHeightLong]{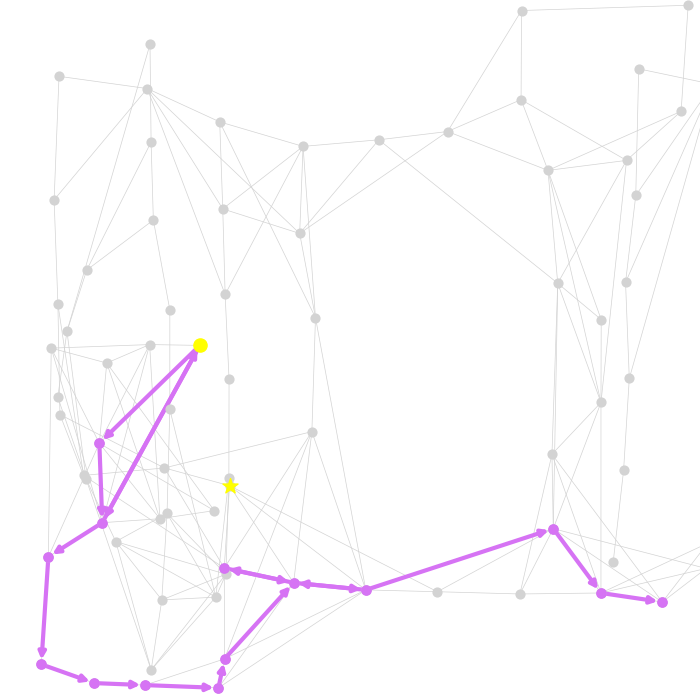} &
\includegraphics[width=\VizImageWidth,height=\VizImageHeightLong]{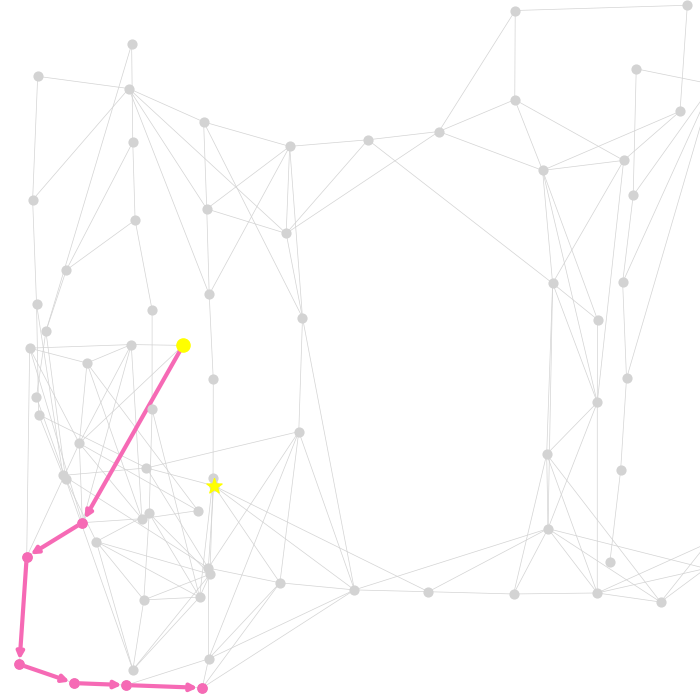} &
\includegraphics[width=\VizImageWidth,height=\VizImageHeightLong]{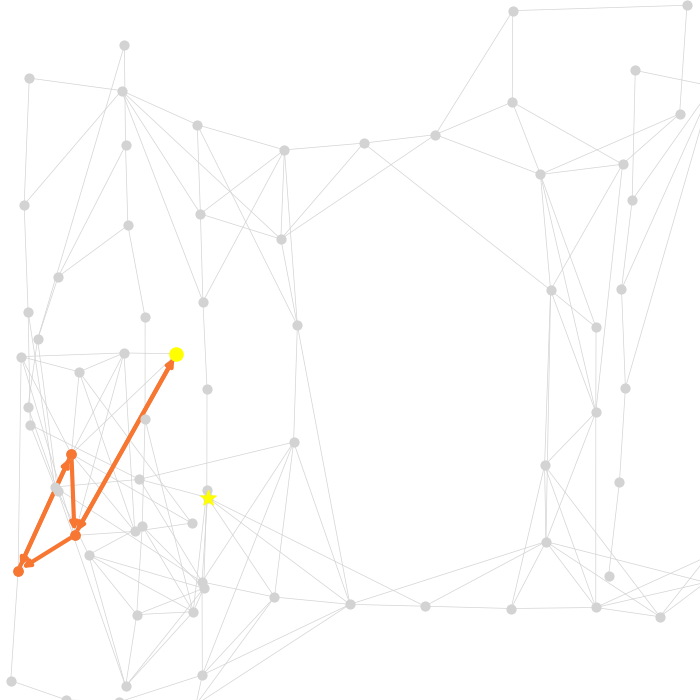} &
\includegraphics[width=\VizImageWidth,height=\VizImageHeightLong]{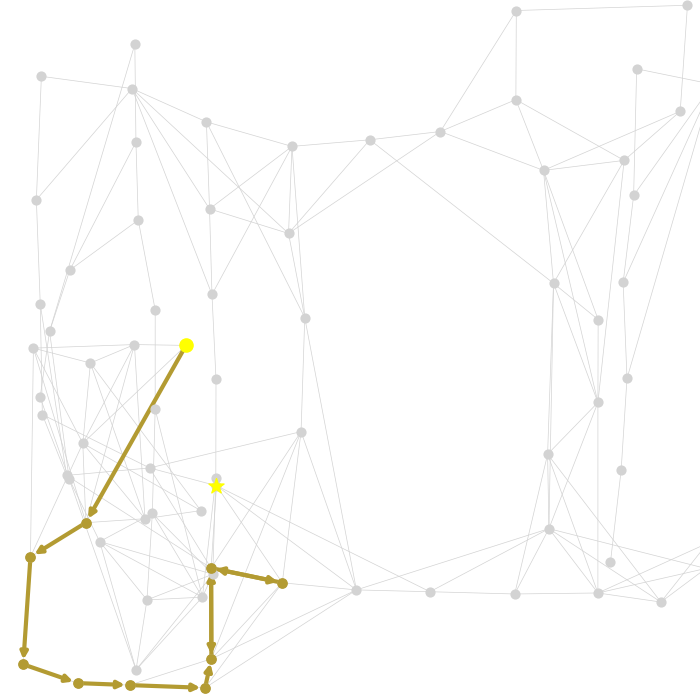} \\

\includegraphics[width=\VizImageWidth,height=\VizImageHeightLong]{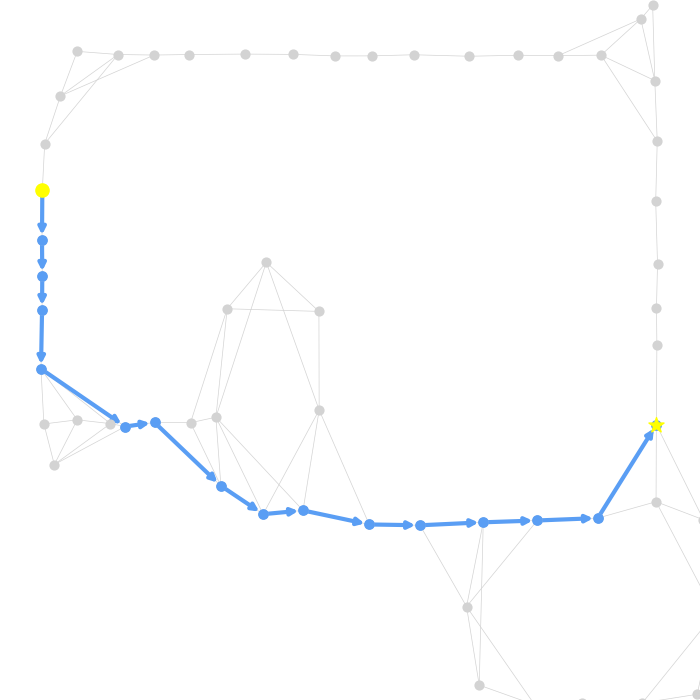} &
\includegraphics[width=\VizImageWidth,height=\VizImageHeightLong]{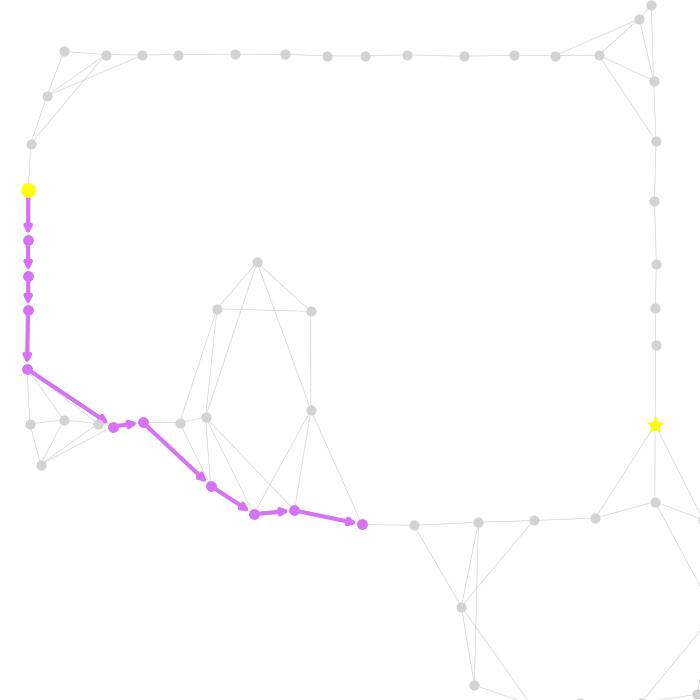} &
\includegraphics[width=\VizImageWidth,height=\VizImageHeightLong]{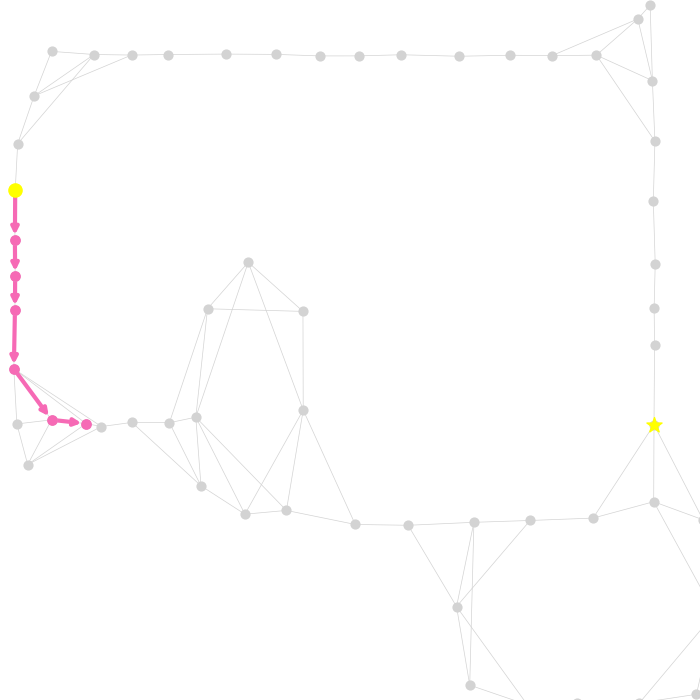} &
\includegraphics[width=\VizImageWidth,height=\VizImageHeightLong]{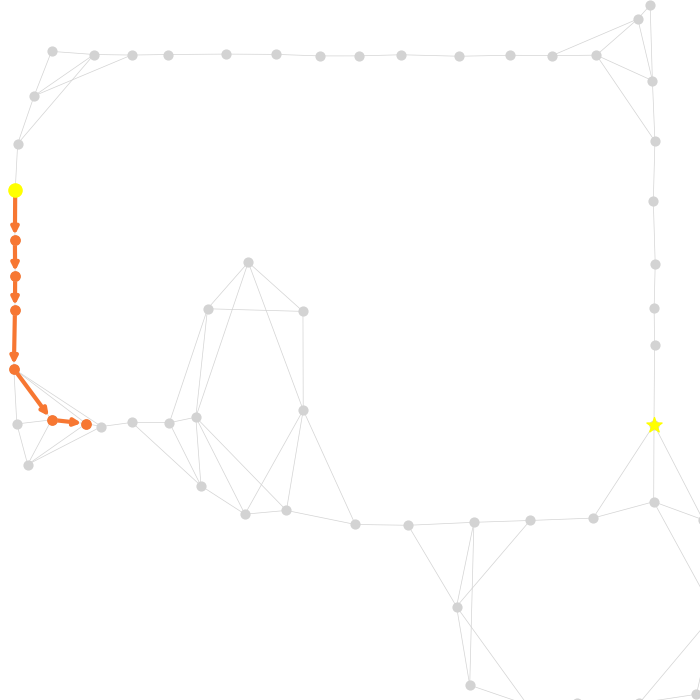} &
\includegraphics[width=\VizImageWidth,height=\VizImageHeightLong]{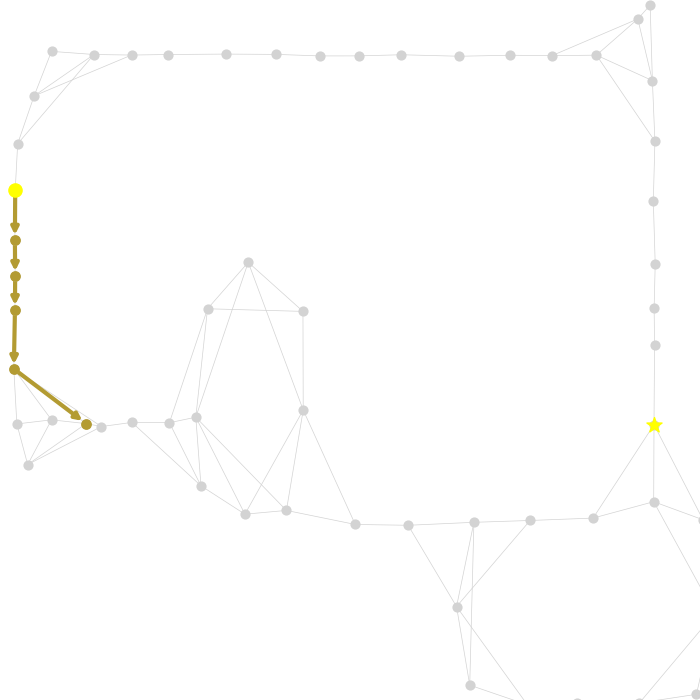} \\

\includegraphics[width=\VizImageWidth,height=\VizImageHeightLong]{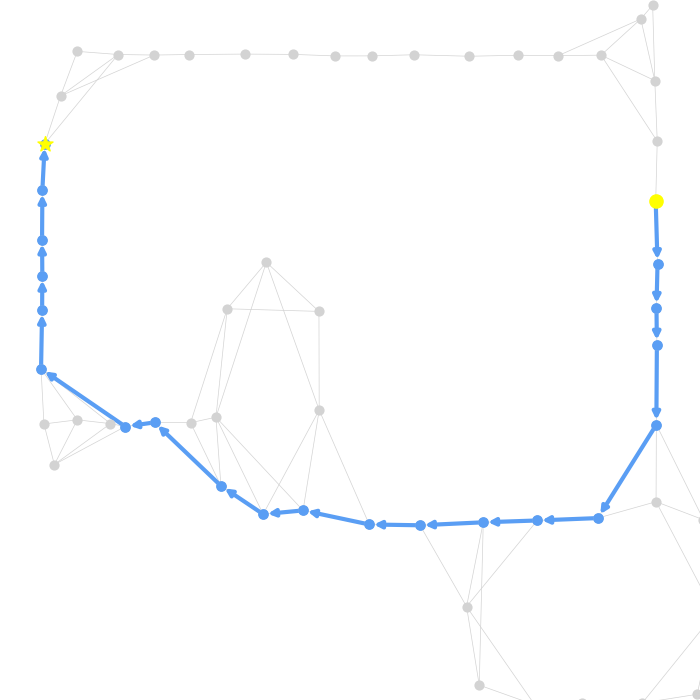} &
\includegraphics[width=\VizImageWidth,height=\VizImageHeightLong]{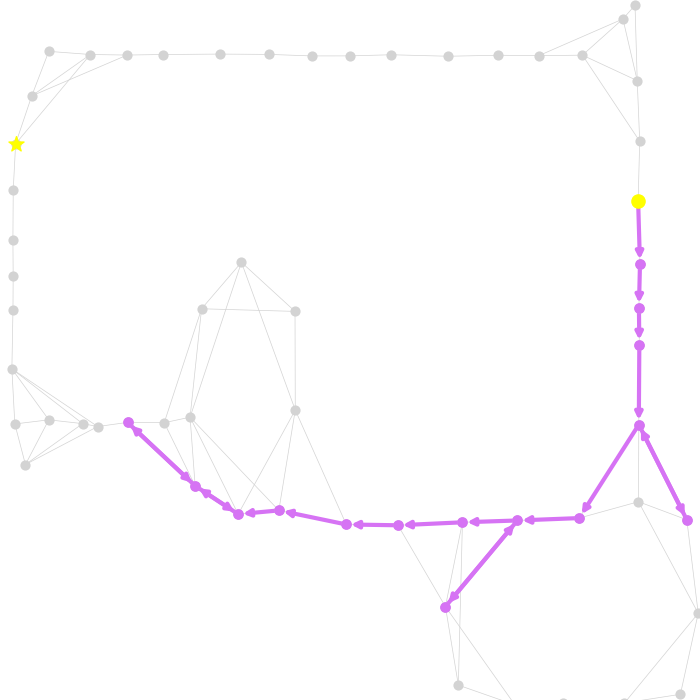} &
\includegraphics[width=\VizImageWidth,height=\VizImageHeightLong]{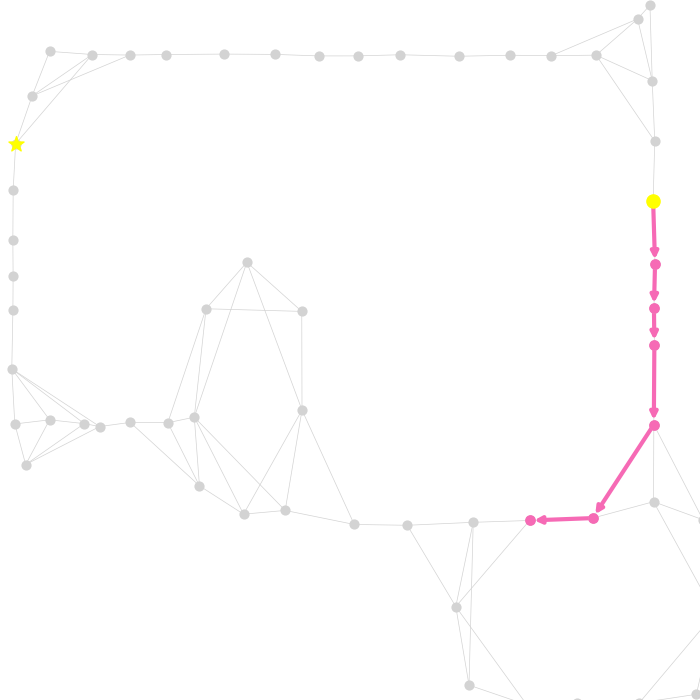} &
\includegraphics[width=\VizImageWidth,height=\VizImageHeightLong]{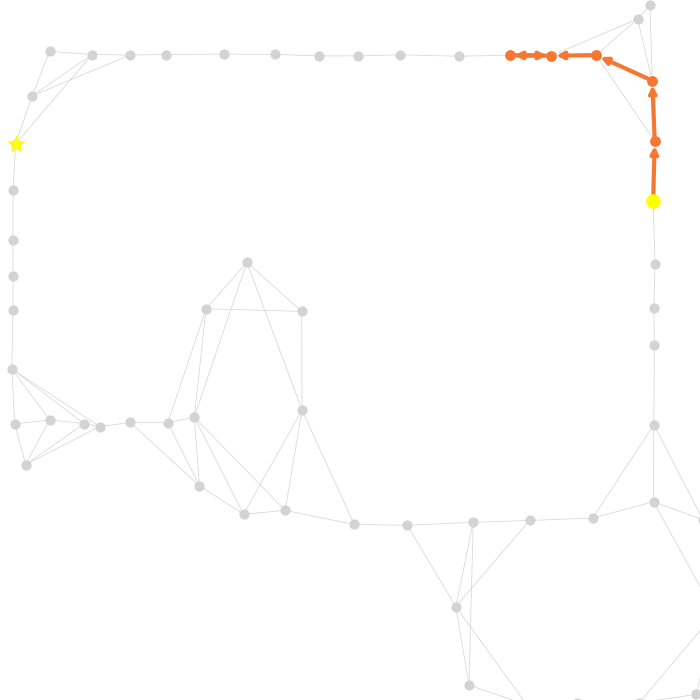} &
\includegraphics[width=\VizImageWidth,height=\VizImageHeightLong]{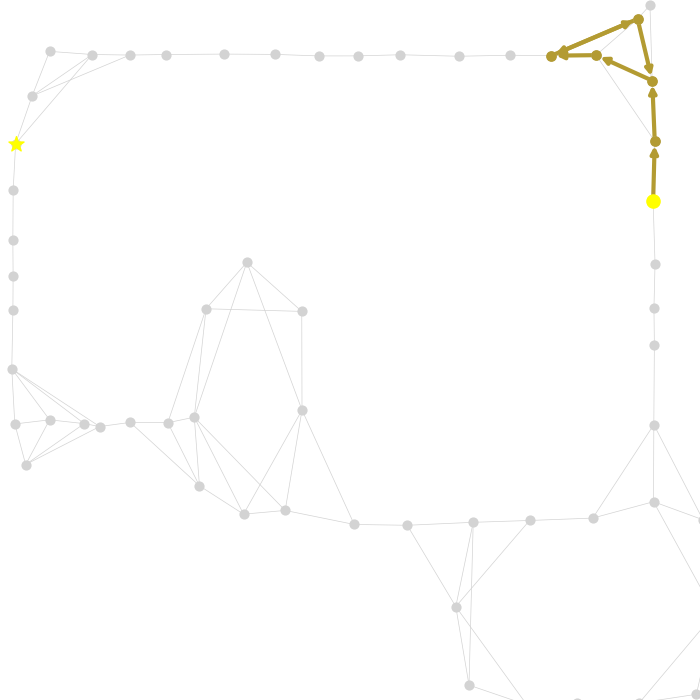} \\

\bottomrule
\end{tabular}
\caption{
    Additional trajectories by human experts and VLN agents on two navigation tasks.
}
\label{fig:sup:viz:path}
\end{figure*}